\documentclass[11pt]{article}
\PassOptionsToPackage{dvipsnames}{xcolor}

\usepackage[preprint]{acl}

\usepackage{times}
\usepackage{latexsym}

\usepackage[T1]{fontenc}

\usepackage[utf8]{inputenc}

\usepackage{microtype}

\usepackage[varqu]{inconsolata}

\usepackage{graphicx}
\usepackage{alltt}
\usepackage{stmaryrd}

\usepackage{listings}
\lstdefinestyle{sqlstyle}{
  basicstyle=\ttfamily\small,
  columns=fullflexible,
  keepspaces=true,
  breaklines=true,
  breakatwhitespace=true,
  frame=none,
  numbers=none,
  xleftmargin=0.5em,
  aboveskip=0.4em,
  belowskip=0.4em,
  lineskip=-1pt
}

\definecolor{darkblue}{rgb}{0, 0, 0.5}
\hypersetup{colorlinks=true, citecolor=darkblue, linkcolor=darkblue, urlcolor=darkblue}

\usepackage{siunitx}
\usepackage{listings}

\usepackage[normalem]{ulem}
\usepackage{comment}
\usepackage{footmisc}
\usepackage{amsmath}
\usepackage{pifont}
\usepackage{tabularx}
\usepackage{subcaption}
\usepackage{caption, booktabs}
\usepackage{algorithm}
\usepackage[noend]{algpseudocode}
\usepackage{amssymb}
\usepackage{setspace}
\usepackage{bbm}
\usepackage{arydshln}
\usepackage{upquote}
\usepackage{fvextra}
\usepackage[htt]{hyphenat}
\usepackage{xparse}

\DefineVerbatimEnvironment{prompt}{Verbatim}{
  breaklines=true,
  breakanywhere=true,
  breaknonspaceingroup=true,
  breaksymbolleft={},
  breaksymbolright={},
  breakindent=0pt,
  fontsize=\scriptsize,
  formatcom=\ttfamily,
}

\makeatletter
\newcommand\footnoteref[1]{\protected@xdef\@thefnmark{\ref{#1}}\@footnotemark}
\makeatother
\newcolumntype{P}[1]{>{\centering\arraybackslash}p{#1}}

\usepackage{xspace}

\usepackage{graphicx}
\usepackage{adjustbox}
\makeatletter
\newcommand{\thickhline}{
\noalign {\ifnum 0=`}\fi \hrule height 1pt
      \futurelet \reserved@a \@xhline
}
\newcolumntype{"}{@{\hskip\tabcolsep\vrule width 1pt\hskip\tabcolsep}}
\makeatother
\usepackage{mathabx}
\usepackage{amsfonts}

\makeatletter
\newcommand*{\blackleq}{
      \mathrel{
            \mathpalette\@blackleq{}
      }
}
\newcommand*{\@blackleq}[2]{
      \vcenter{
            \m@th
            \setbox0=\hbox{$#1\mkern3mu$}
            \setbox2=\hbox{$#1\vcenter{}$}
            \setbox4=\hbox{\raisebox{-\ht2}[.2pt][.2pt]{$#1-$}}
            \hbox{$#1\blacktriangleleft$}
            \nointerlineskip
            \kern\wd0
            \copy4
      }
}
\makeatother
\usepackage{dsfont}

\usepackage{multirow}

\usepackage{mathtools}

\definecolor{my_blue}{RGB}{0,112,192}

\usepackage{placeins}

\usepackage{tikz}
\usetikzlibrary{positioning,arrows.meta,shapes}

\usepackage{fvextra}

\DefineVerbatimEnvironment{WideMinted}{Verbatim}
{breaklines, fontsize=\small, frame=lines, linenos, breakanywhere, xleftmargin=0pt, xrightmargin=0pt}

\usepackage{enumitem}
\usepackage{listings}
\definecolor{my_green}{rgb}{0.1, 0.5, 0.2}
\definecolor{my_pink}{rgb}{0.9, 0.3, 0.7}
\definecolor{my_blue}{rgb}{0.3, 0.5, 0.8}
\definecolor{my_brown}{HTML}{733120}
\definecolor{my_correct}{HTML}{91cb8c}

\newcommand{\ours}[0]{GATE\xspace}
\newcommand{\ourslong}[0]{Grounding After Test from Execution\xspace}

\title{Bootstrapping Semantic Layer from Execution for Text-to-SQL}

\author{\textbf{
Youngwon Lee%
\quad
Jaejin Kim%
\quad
Seung-won Hwang%
\thanks{~Correspondence to:  \href{mailto:seungwonh@snu.ac.kr}{\texttt{seungwonh@snu.ac.kr}}.}
}\\
Seoul National University
}

\begin{document}
\maketitle

\begin{abstract}
Real-world text-to-SQL is often under-specified until user phrases are grounded in how the database stores values.
Prior work attempts to address this by requiring a semantic layer to specify groundings in advance, but such specifications are often incomplete, especially in expert domains where domain-specific conventions are under-documented.
As this leaves multiple grounding hypotheses open for the same SQL part,
we introduce \ours (\ourslong), which bootstraps missing groundings from execution feedback.
\ours keeps grounding hypotheses open while executing the already grounded parts to obtain observations.
Then, only the hypothesis supported by that observation is grounded and stored as a memory entry, recording what was tested and how the open part should be written in SQL.
These entries accumulate into execution-grounded memory, allowing later steps to reuse supported groundings.
Across real-world and controlled benchmarks, \ours consistently improves over strong baselines, demonstrating that execution can serve not only as validation but also as a bootstrapping mechanism for reusable memory in text-to-SQL.\footnote{Our code and data are publicly available at: \url{https://github.com/ludaya/gate}.}
\end{abstract}

\section{Introduction}
\label{sec:intro}

Real-world text-to-SQL often requires grounding underspecified queries in how the database stores values, represents formats, and relates tables.
Prior work addresses this problem by specifying a ``semantic layer'' in advance through documentation, expert annotations, database descriptions~\citep{huang2023documentation,hong2024knowledge,kobayashi-etal-2025-read,shkapenyuk2025metadata}.
However, such specifications are often incomplete, especially in expert domains such as clinical data, where writing the right SQL depends on expert conventions that are under-documented.

\begin{figure*}[t]
\centering
\includegraphics[width=.85\linewidth]{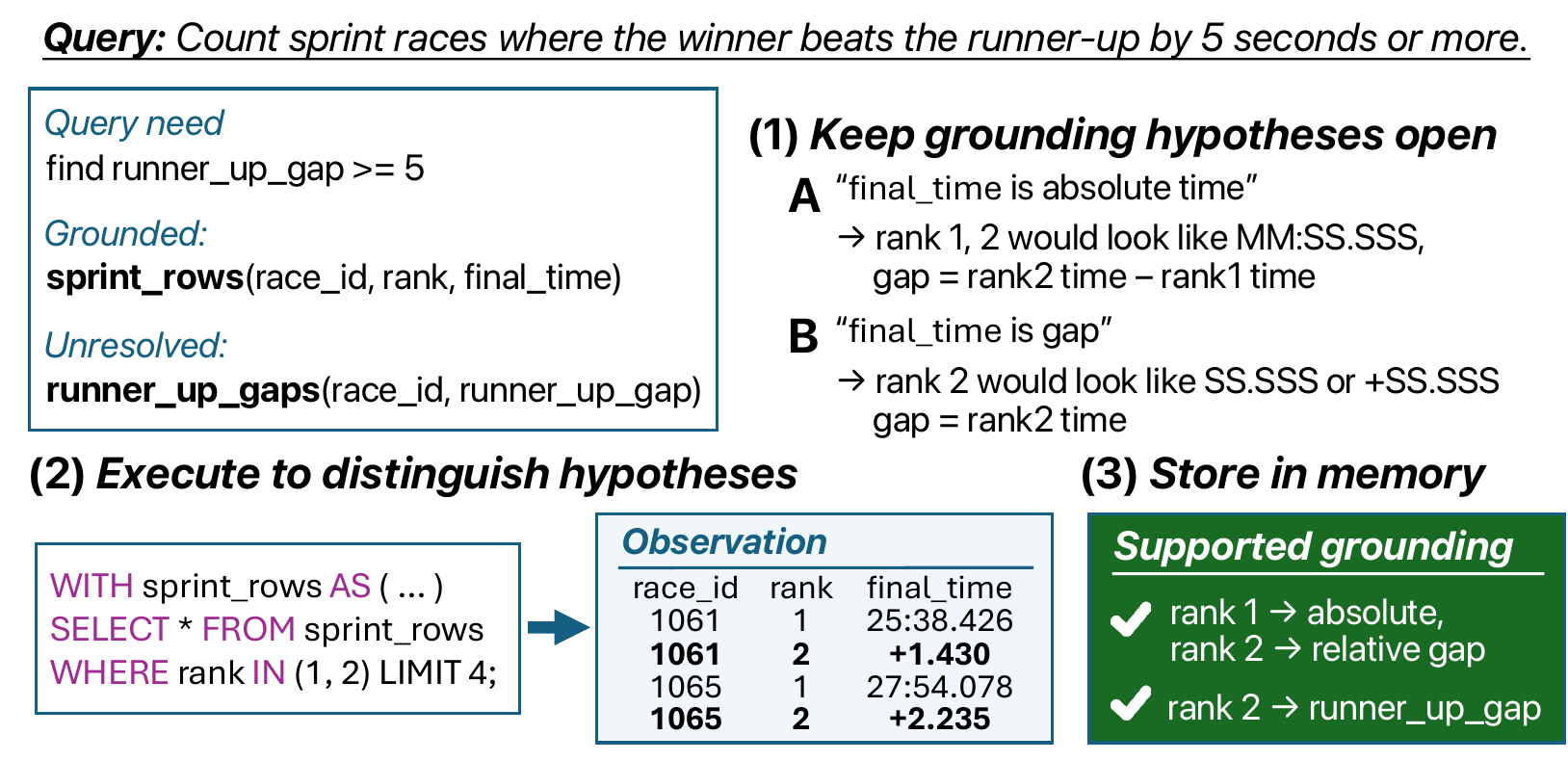}
\caption{
\textbf{Keeping grounding hypotheses open until execution supports one.}
The query requires \texttt{runner\_up\_gap}, but the part producing it remains open while \texttt{sprint\_rows} is already grounded and executable.
Rather than writing one grounding hypothesis into SQL upfront, \ours (1) keeps grounding hypotheses open and (2) executes over \texttt{sprint\_rows}.
(3) The observation supports the grounding that treats rank-2 \texttt{final\_time} as the runner-up's gap, yielding a memory entry for \texttt{runner\_up\_gaps} that can later be reused.
}
\label{fig:partial-plan}
\end{figure*}

Figure~\ref{fig:partial-plan} illustrates this challenge in a query that counts sprint races where the winner beats the runner-up by at least five seconds.
The already grounded part, \texttt{sprint\_rows}, provides \texttt{race\_id}, \texttt{rank}, and \texttt{final\_time}.
What remains open is how to compute \texttt{runner\_up\_gap} from \texttt{final\_time}: \texttt{final\_time} may denote an absolute race time, or it may denote the gap from the winner.
These are \emph{hypotheses} about how \texttt{runner\_up\_gap} should be written in SQL.
Once \texttt{sprint\_rows} exposes rank-1 and rank-2 rows, executing over these rows reveals that rank 1 stores an absolute time such as \texttt{25:38.426}, while rank 2 stores a relative gap such as \texttt{+1.430}.
This observation supports the hypothesis that \texttt{runner\_up\_gap} should be parsed from the rank-2 value.
Once stored, this grounding can be reused by later steps that face the same open part.

In Figure~\ref{fig:partial-plan}, execution is useful because it distinguishes two hypotheses for the same open part.
Prior work typically uses execution after one such hypothesis has already been committed, either by writing it into generated SQL or by fixing a decomposition.
One line writes a possible grounding into SQL and uses execution to filter, repair, or rank candidate queries~\citep{pourreza2024chasesql,lei_spider_2025,borchmann-wydmuch-2025-query,deng_reforce_2025,pham-etal-2026-avsql}.
Another line decomposes the query into smaller problems or planned steps before execution has exposed how the open part should be grounded~\citep{eyal-etal-2023-semantic,pourreza-rafiei-2023-dinsql}.
In both cases, execution may improve the produced SQL, but the hypotheses 
are prematurely collapsed, before execution can test them.

We introduce \ours (\ourslong), which turns execution from post-hoc SQL validation into a bootstrapping mechanism for execution-grounded memory.
\ours treats each open part as a set of grounding hypotheses.
\ours tests a hypothesis through execution once the already grounded parts provide enough context.
If the database observation supports the hypothesis, \ours stores it as a memory entry.
As these entries accumulate, later steps reuse them instead of resolving the same open parts again from scratch.

We evaluate \ours across three representative settings where the semantic layer is incomplete in different ways.
\textbf{RealEHR} is the least-controlled clinical deployment setting,
where the required groundings reflect actual hospital usage, database design, and mismatches with model priors.
\textbf{EHRSQL}~\citep{lee2022ehrsql} retains clinical database challenges while giving each question a shared target SQL.
\textbf{LS-Hard} is the most controlled setting: we derive it from LiveSQLBench~\citep{livesqlbench2025} by withholding the needed groundings, which must instead be recovered through web search.
Across these settings, \ours consistently improves over strong baselines, showing that execution can bootstrap reusable memory for text-to-SQL rather than serving only as final validation.

Our contributions are as follows:
\begin{itemize}
\item We formulate underspecified text-to-SQL as leaving open SQL parts with multiple grounding hypotheses.
\item We propose \ours, which keeps these hypotheses open until execution can distinguish them and stores supported groundings as memory entries.
\item We construct LS-Hard and evaluate \ours across real-world, public clinical, and controlled settings, showing consistent gains over strong baselines.
\end{itemize}

\section{Related Work}
\label{sec:related}
In this section, we review prior text-to-SQL work by when groundings are supplied, written into SQL, or tested.

\paragraph{Pre-specified grounding.}
One line of work supplies grounding in advance by constructing or injecting a semantic layer before inference.
ReGrouP~\citep{dou2023KnowledgeIntensive} and KAT-SQL~\citep{baek-etal-2025-knowledge} transfer domain knowledge across databases, while YORO~\citep{kobayashi-etal-2025-read}, documentation-aware retrieval~\citep{huang2023documentation}, Knowledge-to-SQL~\citep{hong2024knowledge}, and automatic metadata extraction~\citep{shkapenyuk2025metadata} build schema- or database-level descriptions for a target database.
These methods are most effective when the set of needed groundings can be anticipated and remains small enough to specify.

\paragraph{Grounding early, then refining.}
Another line of work writes executable SQL early and uses execution feedback to refine or select among results.
\citet{pham-etal-2026-avsql} construct executable CTE-style views before composing the final SQL, while execution-guided methods use executable queries or prefixes to filter, compare, or select SQL candidates~\citep{borchmann-wydmuch-2025-query}.
Agentic and candidate-based text-to-SQL systems similarly produce concrete SQL attempts or candidates and refine or select among them~\citep{pourreza2024chasesql,lei_spider_2025,deng_reforce_2025}.
While these methods can interact with the database, the executed SQL already has a grounding hypothesis written into it that execution can only accept, reject, or repair post-hoc.

\paragraph{Decomposition for grounding.}
A complementary approach first decomposes the query into subproblems that can be grounded.
\citet{eyal-etal-2023-semantic} introduce a modular query-plan language that decomposes SQL into simpler subqueries, while \citet{pourreza-rafiei-2023-dinsql} decompose text-to-SQL prompting into smaller sub-problems with self-correction.
Delegation-based systems such as ReDel~\citep{zhu2024redel} similarly assign unresolved work to child agents.
While these methods simplify generation, the decomposition is fixed before sufficient database observations are available to ground its parts.

\paragraph{Our distinction.}
Prior work grounds too early: before inference, during execution, or during decomposition.
\ours keeps grounding hypotheses open until execution can distinguish them, then stores the supported grounding in memory for reuse.

\section{Method}
\label{sec:proposed_method}

\ours bootstraps execution-grounded memory by representing open grounding hypotheses as unresolved operators (\S\ref{subsec:plan_formulation}), using execution to distinguish and ground hypotheses (\S\ref{subsec:partial_execution}), and iterating this process while storing and reusing supported groundings (\S\ref{subsec:iterative_search}).

\begin{figure*}[t]
\centering
\includegraphics[width=\textwidth]{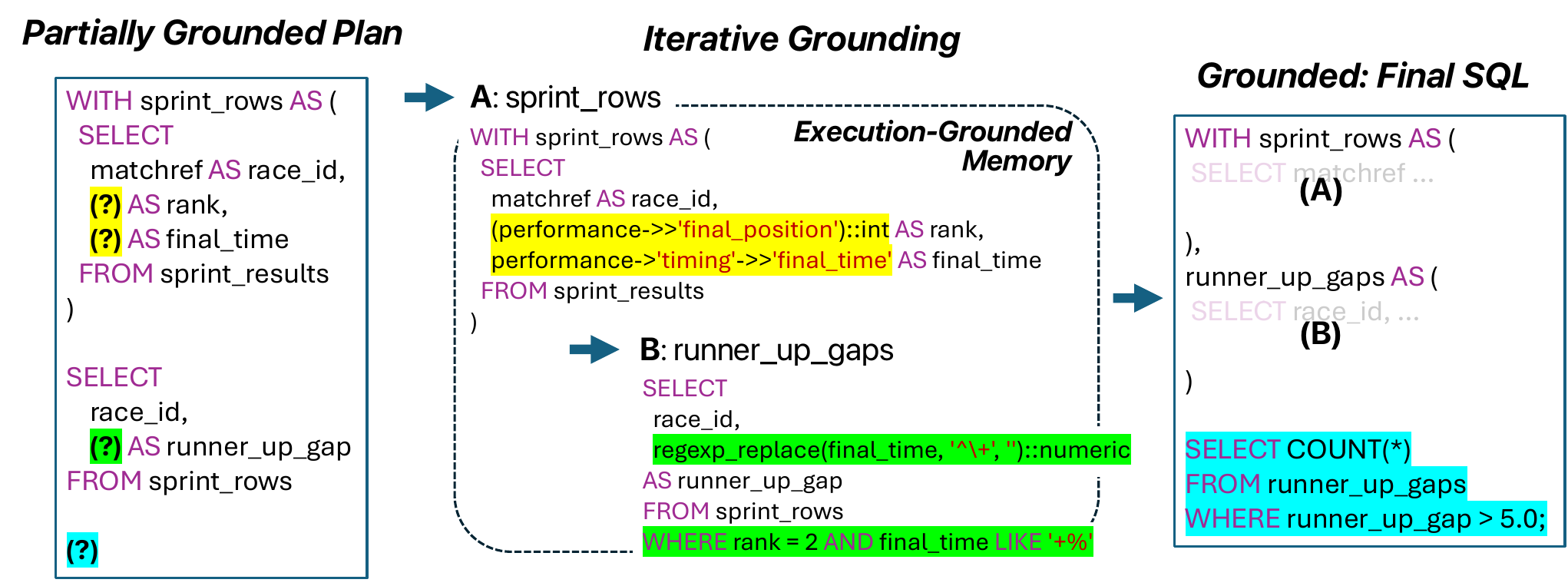}
\caption{
\textbf{Iterative grounding with execution-grounded memory.}
\ours represents the query as view-like operators whose unresolved SQL fragments remain open until execution supports a grounding.
Grounding operator A stores a memory entry for \texttt{sprint\_rows}; grounding operator B adds a memory entry for \texttt{runner\_up\_gaps}.
Repeating this process grows execution-grounded memory with supported groundings, and the accumulated entries are reused to complete the final SQL.
}
\label{fig:plan-update}
\end{figure*}

\subsection{Partially Grounded Plans}
\label{subsec:plan_formulation}

Given a natural-language query \(q\) over a database \(\mathcal{D}\), the goal is to construct the final SQL query \(x^*\).
We formulate SQL construction as completing a \emph{partially grounded plan}: a plan composed of operators, where some are ungrounded.

An operator can be viewed as a CTE-like unit in a \texttt{WITH} clause: a named subquery that materializes part of the computation needed for the final query.
For example, \texttt{sprint\_rows} in Figure~\ref{fig:partial-plan} is such an operator.

We describe each operator \(v\) by its interface and body.
The interface is \((I_v,O_v)\), where \(I_v\) denotes the input dependencies of \(v\), such as base relations or earlier operators that \(v\) may refer to, and \(O_v\) denotes the intended output schema.
The body \(\mathrm{body}(v)\) is the SQL expression that realizes this interface when the operator is grounded.

If the body of \(v\) is already written, we write \(\mathrm{body}(v)=x_v\), where \(x_v\) is a concrete SQL body.
Evaluating \(x_v\) under the database \(\mathcal{D}\), with the dependencies in \(I_v\) available, produces a relation \(R_v\) whose schema matches \(O_v\):
\begin{equation}
    \llbracket x_v \rrbracket_{\mathcal{D}} = R_v,
    \quad
    \mathrm{schema}(R_v)=O_v.
\end{equation}
We call \(v\) \emph{grounded} when this body is written and its dependencies in \(I_v\) are available.
If the body is not yet known, \(v\) is \emph{ungrounded}: its interface \((I_v,O_v)\) is fixed, but the relational mapping from \(I_v\) to \(O_v\) remains unresolved.
The fixed interface lets \ours keep the open part explicit without writing one grounding hypothesis into SQL.
We denote this unresolved mapping by \(\Phi_v\), written schematically as
\begin{equation}
    \Phi_v : I_v \rightarrow O_v .
\end{equation}
A partially grounded plan becomes fully grounded when every unresolved mapping \(\Phi_v\) has been realized by a concrete SQL body.

A partially grounded plan may not be executable as a complete SQL query, but its grounded operators remain executable.
This lets \ours leave unresolved operators open while still executing the operators that are already grounded.
As supported groundings are stored and reused in later iterations, more operators become grounded and can guide subsequent executions.

For example, the operator \texttt{sprint\_rows} in Figure~\ref{fig:partial-plan} is grounded.
It extracts \texttt{race\_id}, \texttt{rank}, and \texttt{final\_time}.
By contrast, the operator \texttt{runner\_up\_gaps} is ungrounded:
\begin{equation}
\begin{aligned}
I_v &= \{\texttt{sprint\_rows}\},\\
O_v &= (\texttt{race\_id},\texttt{runner\_up\_gap}),\\
\Phi_v &= \Phi_{\mathrm{gap}}.
\end{aligned}
\end{equation}
The plan fixes that \texttt{runner\_up\_gap} must be derived from \texttt{sprint\_rows}, but leaves open how \texttt{final\_time} should be used to ground \texttt{runner\_up\_gap}.
For readability, we write this  mapping as
\begin{equation}
    \texttt{runner\_up\_gaps}
    :=
    \Phi_{\mathrm{gap}}(\texttt{sprint\_rows}).
\end{equation}

\subsection{Grounding Through Execution}
\label{subsec:partial_execution}

Given an ungrounded operator \(v\) in a partially grounded plan \(P\), 
\ours proposes grounding hypotheses, derives an executable SQL query over the already-grounded operators, and uses the observation to decide how \(v\) should be grounded.

\paragraph{Generating grounding hypotheses.}
An ungrounded operator can be grounded in multiple ways.
We represent each grounding hypothesis as
\begin{equation}
h_k=(\Phi^{(k)}_v,E_k),
\label{eq:candidate_grounding}
\end{equation}
where \(\Phi^{(k)}_v\) is a candidate mapping consistent with the interface \((I_v,O_v)\), and \(E_k\) is a predicate describing what observation would support that mapping.
For the runner-up example, let \(v=\texttt{runner\_up\_gaps}\), let \(S=\texttt{sprint\_rows}\), let \(W=\sigma_{r=1}(S)\) be the winner rows, and let \(R=\sigma_{r=2}(S)\) be the runner-up rows.
Two possible mappings are:
\begin{equation}
\begin{aligned}
\Phi^{(1)}_v(S)
&=
\pi_{id,\; \tau(R.t)-\tau(W.t)\rightarrow gap}
   (R \bowtie_{id} W),\\
\Phi^{(2)}_v(S)
&=
\pi_{id,\; \rho(t)\rightarrow gap}(R).
\end{aligned}
\end{equation}
Here, \(id\), \(r\), and \(t\) abbreviate \texttt{race\_id}, \texttt{rank}, and \texttt{final\_time}; \(\tau\) parses an absolute race time, and \(\rho\) parses a relative gap.
The first grounding subtracts the winner's time from the runner-up's time, while the second parses the rank-2 value directly as the gap.
The predicates \(E_k\) specify how these hypotheses should appear in an observation: the first expects rank-1 and rank-2 \texttt{final\_time} values to follow an absolute-time pattern, while the second expects the rank-2 value to follow a relative-gap pattern.

\paragraph{Testing with grounded operators.}
Rather than committing to one hypothesis, \ours derives an executable SQL query \(\tilde{x}_v\) over the already grounded operators.
This query does not require \(v\) to be written as SQL; it gathers an observation that distinguishes the grounding hypotheses.
The database returns the observation
\begin{equation}
o = \llbracket \tilde{x}_v \rrbracket_{\mathcal{D}}.
\label{eq:grounded_observation}
\end{equation}
An observation supports hypothesis \(h_k=(\Phi^{(k)}_v,E_k)\) when
\begin{equation}
    o \models E_k .
\end{equation}

\paragraph{Grounding the operator.}
After observing \(o\), \ours resolves the unresolved mapping:
\begin{equation}
    \Phi_v \leadsto \hat{\Phi}_v .
\end{equation}
Concretely, the LLM compares \(o\) with the predicates \(E_1,\ldots,E_m\), identifies the grounding supported by the observation, and returns both the resolved mapping and its SQL realization:
\begin{multline}
(\hat{\Phi}_v,\, \hat{x}_v)
= \\
m_{\mathrm{LLM}} \left(
    q,\,
    P,\,
    v,\,
    \{(\Phi^{(k)}_v,E_k)\}_{k=1}^{m},\,
    \tilde{x}_v,\,
    o
\right).
\label{eq:operator_grounding}
\end{multline}
For example, observing \texttt{+1.430} supports treating rank-2 \texttt{final\_time} as a relative gap and shows that the SQL parser should handle a leading \texttt{+}.
The returned \(\hat{x}_v\) grounds \(v\) in the updated plan:
\begin{equation}
    \Phi_v \leadsto \hat{\Phi}_v,
    \quad
    \mathrm{body}_{P^+}(v)=\hat{x}_v .
\end{equation}
In the runner-up example, the SQL realization is:
\begin{lstlisting}[style=sqlstyle, language=SQL]
SELECT race_id,
       regexp_replace(final_time, '^\+', '')::numeric AS runner_up_gap
FROM sprint_rows
WHERE rank = 2 AND final_time LIKE '+%';
\end{lstlisting}
Thus, \texttt{runner\_up\_gaps} becomes grounded by deriving \texttt{runner\_up\_gap} from rank-2 \texttt{final\_time}.

If no grounding hypothesis is supported, \ours executes another SQL query for the same operator or revises the plan structure.
One observation may resolve only part of the grounding: for example, one query may establish that rank-2 \texttt{final\_time} stores a relative gap, while a later query reveals that some gaps use \texttt{+M:SS.SSS} rather than \texttt{+S.SSS}.
Such observations specialize the grounding without restarting the plan.

\subsection{Iterative Grounding and Reuse}
\label{subsec:iterative_search}

\ours repeats operator grounding described in \S\ref{subsec:partial_execution} 
until all unresolved operators are grounded.
Figure~\ref{fig:plan-update} illustrates this iterative process.
A partially grounded plan may contain several unresolved operators, shown as colored holes in the SQL.
For example, the yellow hole is filled with a concrete SQL body that grounds \texttt{sprint\_rows}, and the supported grounding is stored in memory as entry A for later reuse.
Similarly, the green hole is filled with the SQL body that grounds \texttt{runner\_up\_gaps}, and its supported grounding is stored as entry B.
The final SQL is completed by reusing entries A and B.
More generally, each iteration uses execution feedback to fill one open part and store the supported grounding, so plan completion and memory accumulation proceed together.

This raises the next question: when multiple operators remain ungrounded, which one should \ours ground next?

\paragraph{Plan evaluation and selection.}
At each iteration, \ours must evaluate where each partial plan is blocked and select which unresolved operator to ground next, because grounding decisions have dependencies.
For example, \texttt{runner\_up\_gaps} can be tested only after \texttt{sprint\_rows} has exposed \texttt{race\_id}, \texttt{rank}, and \texttt{final\_time}.
Rather than following a single refinement chain, \ours maintains multiple candidate partially grounded plans, each corresponding to a different sequence of grounding decisions.

For evaluation, \ours scores these plans along three dimensions.
\emph{Progress} measures how much of the final query has already been grounded.
\emph{Bottleneck} identifies the location of the current blockage: the unresolved operator whose grounding most limits executability or downstream progress.
\emph{Potential} estimates how much benefit would be gained by resolving that bottleneck.
Together, these assessments estimate which candidate plans are most promising for further refinement.
Based on this evaluation, \ours selects the candidate plan whose bottleneck is both currently testable and most informative given the operators already grounded.

\paragraph{Summarizing execution evidence.}
Grounding a bottleneck operator may require more than one execution, because a single observation can establish, refine, or rule out only part of a candidate grounding.
\ours therefore summarizes the observations obtained from executing grounded subplans and conditions later executions for the same bottleneck on this summary.
This lets successive grounding attempts build on prior evidence rather than treating each attempt as independent.

\paragraph{Reusing grounding updates.}
When execution supports a grounding for an operator v, \ours stores the result in execution-grounded memory as a grounding update.
A grounding update is a memory entry that records the executable query, the observation, and the SQL body that realizes the supported mapping.
Because the update is stored outside any single partially grounded plan, it can be applied to another partially grounded plan when that plan contains an unresolved operator with the same interface and the update does not conflict with already grounded operators.
This cross-plan reuse allows a grounding established in one partially grounded plan to constrain another plan without repeating the same database interaction.

\paragraph{Completion.}
If all operators in a partially grounded plan are grounded, the plan yields an executable SQL query, which is added to the candidate set \(\mathbb{C}\).
Search terminates when enough candidates have been gathered or the iteration budget is exhausted, and an LLM judge selects the final answer from \(\mathbb{C}\) after execution-based checking.
Algorithm~\ref{alg:ours} in App.~\ref{app:algorithm} gives the implementation details: candidate plans are represented as trajectories of grounding steps, and the algorithm specifies plan evaluation, grounding-update memory, and archive updates.

\section{Experiments}
\label{sec:exp}

We evaluate \ours along three questions:
\begin{itemize}
\item Does \ours improve text-to-SQL performance when database-specific grounding must be recovered during inference? (\S\ref{subsec:results})
\item Which components make iterative grounding effective, and are the gains more than simply making more attempts?
(\S\ref{subsec:ablation}, \S\ref{subsec:cost_analysis})
\item How much reusable memory does \ours accumulate, and are the stored grounding updates correct and consistent?
(\S\ref{subsec:update_reliability})
\end{itemize}

\begin{table*}[t]
\centering
\resizebox{\textwidth}{!}{%
\setlength{\tabcolsep}{4pt}
\begin{tabular}{l cc c cc c}
    \toprule
    & \multicolumn{3}{c}{\textbf{Qwen3-8B}}
    & \multicolumn{3}{c}{\textbf{GPT-5.4-mini}} \\
    \cmidrule(lr){2-4} \cmidrule(lr){5-7}
    \textbf{Method}
    & RealEHR & EHRSQL & LS-Hard
    & RealEHR & EHRSQL & LS-Hard \\
    \midrule
    ReAct~\citep{yao_react_2023}
     & 1.7 & 30.2 & 14.5
     & 42.9 & 32.2 & 37.8 \\
    ReDel~\citep{zhu2024redel}
     & 2.5 & 26.4 & 8.0
     & 44.8 & 26.8 & 40.2 \\
    ReFoRCE~\citep{deng_reforce_2025}
    & 3.8 & 31.2 & 13.9
    & 48.6 & 27.6 & 40.1 \\
    \textbf{\ours} (proposed)
     & \textbf{5.6} & \textbf{35.0} & \textbf{17.3}
     & \textbf{55.2} & \textbf{34.4} & \textbf{45.1} \\ 
    \bottomrule
\end{tabular}}
\caption{
Performance across three benchmarks grouped by backbone LLM.
RealEHR and LS-Hard report answer accuracy (Acc, \%); EHRSQL reports execution accuracy (EX, \%).
Standard deviations across runs are reported in App.~\ref{app:run_variance}.
}
\label{tab:main_results}
\end{table*}

\subsection{Experimental Settings}
\label{subsec:exp_settings}

\paragraph{Benchmarks and metrics.}
We evaluate \ours on three benchmarks that span different degrees of control over missing grounding (details in App.~\ref{app:benchmark_details}).
RealEHR ($n{=}47$) is the least controlled setting: it is a closed clinical benchmark collected from actual hospital data, where the required groundings reflect institution-specific conventions that vary across queries; queries and answers are curated by medical professionals (examples in App.~\ref{app:realehr_examples}).
EHRSQL~\citep{lee2022ehrsql} ($n{=}100$) is a public clinical benchmark built on MIMIC~\citep{johnson2023mimic}; each question has a fixed target SQL, while retaining grounding challenges induced by clinical database conventions (subset construction in App.~\ref{app:ehrsql_subset}).
LS-Hard ($n{=}180$) is the most controlled setting: it is derived from LiveSQLBench~\citep{livesqlbench2025} by removing the provided statements that make the needed groundings explicit; those groundings must instead be recovered through web search.
We report answer accuracy on RealEHR and LS-Hard using an LLM judge (App.~\ref{app:llm_judge}), and execution accuracy (EX) on EHRSQL.

\paragraph{Baselines.}
We compare \ours against three representative baselines that instantiate the main alternatives discussed in Section~\ref{sec:related}.
ReAct~\citep{yao_react_2023} represents methods that interact with the database through executable SQL attempts and use the resulting feedback to guide later steps.
ReFoRCE~\citep{deng_reforce_2025} is a stronger executable-attempt baseline that adds self-refinement, format restriction, and column exploration.
ReDel~\citep{zhu2024redel} represents the decomposition-first alternative, where subtasks are delegated to separate agents before their results are integrated.
More implementation details are provided in App.~\ref{app:impl_details}, along with prompt templates in App.~\ref{app:prompts}.

\subsection{Main Results}
\label{subsec:results}

Table~\ref{tab:main_results} reports results across the three benchmarks.
\ours achieves the best performance across all settings, showing that grounding unresolved operators only when they become testable is effective across different forms of missing grounding.
The results provide complementary evidence across the evaluation spectrum: RealEHR tests the least controlled clinical deployment setting, EHRSQL tests a fixed-target public clinical setting, and LS-Hard tests controlled missing grounding.
We also provide a detailed trace comparison in App.~\ref{app:qual_analysis} showing how ReAct commits to an incorrect grounding and fails to revise it, whereas \ours identifies the remaining bottleneck, executes a follow-up query, and stores the supported grounding.

\paragraph{RealEHR: real-world clinical deployment setting.}
On RealEHR, \ours achieves the best accuracy under both backbone LLMs, reaching 55.2\% with GPT-5.4-mini and 5.6\% with Qwen3-8B.
RealEHR is the least controlled setting in our evaluation: clinical shorthand, institution-specific codes, heterogeneous value names, and underdocumented database conventions can all introduce missing grounding, and the query itself may leave residual ambiguity.
The result shows that \ours can recover the grounding needed for the current query even when multiple sources of uncertainty coexist, not only when the missing grounding is controlled by benchmark construction.

\paragraph{EHRSQL: fixed-target public clinical setting.}
On EHRSQL, \ours again achieves the best execution accuracy under both backbone LLMs, reaching 34.4\% EX with GPT-5.4-mini and 35.0\% with Qwen3-8B.
EHRSQL provides a public clinical database and a fixed target SQL for each question, giving a more controlled clinical evaluation setting while still requiring the system to ground questions in database-specific conventions such as time formats, paraphrases, and shifted temporal values.
The result shows that our gains are not limited to RealEHR's institution-specific, less-controlled deployment setting.

\paragraph{LS-Hard: controlled missing-grounding setting.}
On LS-Hard, \ours achieves the best accuracy under both backbone LLMs, reaching 45.1\% with GPT-5.4-mini and 17.3\% with Qwen3-8B.
LS-Hard provides the most controlled test of our mechanism.
We withhold selected information that would otherwise make the needed grounding explicit, so the system must distinguish grounding hypotheses during inference.
At the same time, the withheld information can be recovered through web search.
The result shows that \ours improves inference-time grounding recovery beyond clinical-domain-  or benchmark-specific priors.

\begin{table}[t]
    \centering
    \setlength{\tabcolsep}{3pt}
    \begin{tabular}{l c}
        \toprule
        \textbf{Variant} & \textbf{RealEHR} \\
        \midrule
        \ours (Full)
         & \textbf{55.2} \\
        \phantom{0}$-$ Plan Evaluation
         & 44.7 \\
         \phantom{0}$-$ Plan Selection
         & 53.2 \\
        \phantom{0}$-$ Evidence Summarization
         & 46.8 \\
        \phantom{0}$-$ Grounding Update Reuse
         & 53.2 \\
        \bottomrule
    \end{tabular}
    \caption{
Ablation study on RealEHR:
Results show answer accuracy (\%) from GPT-5.4-mini.
    }
    \label{tab:ablation}
\end{table}

\begin{table}[t]
    \centering
    \setlength{\tabcolsep}{4pt}
    \begin{tabular}{l cc}
        \toprule
        \textbf{Method}
         & \textbf{Acc}
         & \textbf{Cost ($\downarrow$)} \\
        \midrule
        ReAct
         & 42.9
         & 6.3 \\
        ReAct, 6 attempts
         & 51.1
         & 39.1 \\
        \textbf{\ours}
         & \textbf{55.2}
         & 36.0 \\
        \bottomrule
    \end{tabular}
    \caption{
    Cost analysis with GPT-5.4-mini on RealEHR.
    Cost shows the average number of LLM calls per query.
    ReAct single-run accuracy is the mean of 6 independent runs.
    At comparable budgets, \ours outperforms ReAct $\times 6$ with less cost.
    }
    \label{tab:cost_analysis}
\end{table}

\subsection{Ablation}
\label{subsec:ablation}

To understand how each component in \ours contributes to performance,
we conduct an ablation study on RealEHR using GPT-5.4-mini.
Each row in Table~\ref{tab:ablation} removes one component from the iterative grounding framework:
plan evaluation, plan selection, evidence summarization, or grounding update reuse.

\paragraph{Plan evaluation is critical for identifying meaningful bottlenecks.}
Removing plan evaluation (``$-$ Plan Evaluation'') causes a 10.5pp drop, the largest among the ablations.
Plan evaluation scores candidate partially grounded plans by progress, current bottleneck, and the potential benefit of resolving that bottleneck.
Without this structured evaluation, execution outcomes are not tied to the bottleneck they address or to whether resolving that bottleneck represents genuine progress toward the final query.
This shows that bottleneck identification must be modeled explicitly rather than inferred implicitly from local execution feedback.

\paragraph{Plan selection determines where grounding effort is applied.}
Removing plan selection (``$-$ Plan Selection'') causes a 2.0pp drop.
Plan selection aims to choose the plan whose bottleneck is both currently testable and most useful to resolve next.
The result suggests that iterative grounding benefits from prioritizing which bottleneck to resolve first.

\paragraph{Evidence summarization enables iterative refinement of bottleneck grounding.}

Removing evidence summarization (``$-$ Evidence Summarization'') causes an 8.4pp drop.
Without summarizing prior observations, repeated execution for the same bottleneck operator becomes a sequence of independent retries.
In contrast, \ours makes refinement cumulative to avoid revisiting failed hypotheses: evidence summaries record what earlier executions established, ruled out, or partially supported, allowing later grounding attempts to focus on the remaining uncertainty.
A concrete example of summaries can be found in App.~\ref{app:qual_analysis}.

\paragraph{Grounding updates generalize across plans.}
Removing update reuse (``$-$ Grounding Update Reuse'') causes a 2.0pp drop.
When a grounding succeeds, the corresponding update records the SQL body and evidence summary.
Reusing this update across compatible partially grounded plans prevents the system from resolving the same operator again.
Without reuse, grounding progress cannot be shared across plans.
Section~\ref{subsec:update_reliability} discusses further details on how many updates are added to the memory and reused.

\subsection{Cost Analysis}
\label{subsec:cost_analysis}

In this section, we compare \ours against repeated ReAct runs under a matched LLM-call budget on RealEHR with GPT-5.4-mini.
For repeated ReAct, we run the agent independently multiple times and use the same backbone LLM to select the final answer; the selector prompt is provided in App.~\ref{app:llm_answer_selector}.

Table~\ref{tab:cost_analysis} shows that a single ReAct run uses 6.3 calls on average and achieves 42.9\% accuracy.
With six times that budget, including one additional call for candidate selection, repeated ReAct reaches 51.1\% at 39.1 calls.
\ours reaches 55.2\% using 36.0 calls, outperforming repeated ReAct by +4.1pp while using fewer calls.

These results show that \ours does not improve merely by making more attempts.
Repeated ReAct runs spend a similar number of calls, but their observations remain isolated within independent trajectories.
By contrast, \ours stores execution results as grounding updates, reuses them across partially grounded plans, and selects the next trajectory based on which grounding decision is most useful to test.
The advantage therefore comes from turning additional calls into structured search progress, rather than from increasing the call budget alone.

\subsection{Update Consistency and Scale}
\label{subsec:update_reliability}

\begin{table}[t]
\centering
\setlength{\tabcolsep}{4pt}
\begin{tabular}{l r}
    \toprule
    \textbf{Metric} & \textbf{Value} \\
    \midrule
    Grounding updates & 912 \\
    Avg. updates / question & 19.40 \\
    Updates reused across plans & 97.8\% \\
    \midrule
    All possible update pairs & 8,894 \\
    Table-overlap pairs & 6,910 \\
    LLM-flagged contradiction pairs & 32 \\
    Genuine contradiction pairs & 1 \\
    Incorrect update summaries & 5 / 912 \\
    \midrule
    Update accuracy & 99.5\% \\
    Contradiction rate & .011\% \\
    \bottomrule
\end{tabular}
\caption{
Consistency of grounding updates on RealEHR with gpt-5.4-mini backbone.
}
\label{tab:update_reliability}
\end{table}

Because \ours accumulates and reuses grounding updates across partially grounded plans, we audit their scale, correctness, and consistency.
We analyze 912 grounding updates from the 47 RealEHR questions, averaging 19.40 updates per question.
Among these updates, 97.8\% were reused by at least one compatible plan, with 6.4 reuse applications on average.

To assess consistency, we compare pairs of updates within the same question.
We only judge pairs whose executable queries access at least one common table, since updates over disjoint tables are unlikely to directly conflict; this retains 6,910 of 8,894 possible pairs (77.7\%).
GPT-5.4 judges whether each pair is contradictory, where contradiction means that two updates cannot both be true under the same task scope.

Table~\ref{tab:update_reliability} shows that grounding updates are frequent, mostly correct, and mutually consistent.
Manual inspection finds only 5 incorrect update summaries among 912 updates, yielding 99.5\% update accuracy.
The judge initially flags 32 of 6,910 table-overlap pairs as contradictions (0.46\%), but manual review shows that 31 are not genuine conflicts between two valid grounding updates; for those cases, one of the updates was incorrect.
After removing such summary errors, only one genuine conflict remains, corresponding to 0.014\% of judged pairs and 0.011\% of all possible within-question pairs.

The single genuine conflict is also informative.
It arises when the natural-language concept ``eosinophil'' is grounded to two plausible lab measurements, \texttt{Eosinophil} and \texttt{Blood Eosinophil count}, which imply different peak dates and medication evidence.
App.~\ref{app:eosinophil_conflict} analyzes this case in detail.
Overall, the audit suggests that reusable grounding updates are consistent in the least-controlled benchmark setting: they are usually supported by their own observations, and genuine conflicts between valid updates are extremely rare.

\section{Conclusion}
\label{sec:conclusion}

Real-world text-to-SQL often requires grounding underspecified queries in how a database stores values, represents formats, and relates tables.
\ours addresses this setting by keeping grounding hypotheses open until execution can distinguish them, then storing supported groundings as execution-grounded memory.
Across real-world, fixed-target, and controlled settings, \ours consistently improves over strong baselines, showing that execution can bootstrap reusable memory rather than serve only as final validation.

\section*{Limitations}
\label{sec:limitations}
First, \ours assumes that the missing grounding can be recovered from database contents, executable checks, or provided external sources.
Some conventions may remain implicit, undocumented, or inaccessible, and therefore cannot be discovered.

Second, we focus on single-turn query resolution where the intended target is fixed once database-specific grounding is recovered.
Cases where the user's intent is itself underspecified and must be resolved through clarification are outside our scope and left to future work.

\typeout{}
\bibliography{custom,anthology_part1,anthology_part2,anthology_part3,anthology_part4}

\appendix

\begin{algorithm*}[!t]
\caption{\ours: Iterative grounding with execution-grounded memory}
\label{alg:ours}
\begin{algorithmic}[1]
\Require Query \(q\), database \(\mathcal{D}\), iteration budget \(N\), grounding attempts \(J\), min/max candidates \(c_{\min},c_{\max}\)
\Ensure Target SQL \(x^*\)

\State \(\mathbb{T} \gets \{\epsilon\}\) \Comment{Trajectory archive; each trajectory induces a partially grounded plan}
\State \(\mathbb{M} \gets \emptyset\) \Comment{Execution-grounded memory of grounding updates}
\State \(\mathbb{C} \gets \emptyset\) \Comment{Candidate final SQL queries}

\For{\(t \gets 1\) \textbf{to} \(N\)}
    \State \(\mathbb{E}_t \gets \operatorname{EvaluatePlans}(q,\mathbb{T},\mathbb{M})\) \Comment{Evaluate progress, bottleneck, and potential}
    \State \(\hat{T}_t \gets \operatorname{SelectPlan}(q,\mathbb{E}_t)\) \Comment{Select a plan to continue}
    \State \(\hat{P}_t \gets \operatorname{Plan}(\hat{T}_t,\mathbb{M})\)
    \State \(\hat{v}_t \gets \operatorname{Bottleneck}(\hat{T}_t,\mathbb{E}_t)\)

    \State \(B \gets \emptyset\); \quad \(\Delta\mathbb{M} \gets \emptyset\) \Comment{Local evidence summary and new memory entries}

    \For{\(j \gets 1\) \textbf{to} \(J\)}
        \State \((H_j,\tilde{x}_j) \gets \operatorname{ProposeGrounding}(q,\hat{P}_t,\hat{v}_t,B)\) \Comment{Grounding hypotheses and executable SQL query}
        \State \(o_j \gets \operatorname{Exec}(\tilde{x}_j,\mathcal{D})\)
        \State \(u_j \gets \operatorname{SummarizeAndGround}(q,\hat{P}_t,\hat{v}_t,H_j,\tilde{x}_j,o_j,B)\)
        \State \(B \gets \operatorname{UpdateLocal}(B,\tilde{x}_j,o_j,u_j)\)

        \If{\(u_j\) contains a supported grounding for \(\hat{v}_t\)}
            \State \(\Delta\mathbb{M} \gets \Delta\mathbb{M}\cup\{u_j\}\)
            \If{\(\operatorname{Apply}(\hat{P}_t,u_j)\) yields a final SQL query \(x\)}
                \State \(\mathbb{C} \gets \mathbb{C}\cup\{x\}\)
            \EndIf
        \EndIf
    \EndFor

    \State \(\mathbb{M} \gets \mathbb{M}\cup\Delta\mathbb{M}\) \Comment{Add supported groundings to memory}
    \State \(\mathbb{T} \gets \operatorname{UpdateArchive}(\mathbb{T},\hat{T}_t,\Delta\mathbb{M},\mathbb{M})\) \Comment{Append new updates and reuse them on compatible plans}

    \If{\(|\mathbb{C}|\ge c_{\max}\)}
        \State \textbf{break}
    \EndIf
    \If{\(|\mathbb{C}|\ge c_{\min}\) \textbf{and} \(\operatorname{Terminate}(q,\mathbb{C},\mathbb{M})\)}
        \State \textbf{break}
    \EndIf
\EndFor

\State \Return \(\operatorname{SelectAnswer}(q,\mathbb{C},\mathbb{M})\)
\end{algorithmic}
\end{algorithm*}

\FloatBarrier

\section{Full Algorithm}
\label{app:algorithm}

For completeness, we provide the full pseudocode of \ours in this appendix.
Alg.~\ref{alg:ours} summarizes the loop described in \S\ref{sec:proposed_method}: selecting a trajectory that induces a partially grounded plan, executing grounded subplans to ground one unresolved operator, storing the resulting execution step as a grounding update, and updating the trajectory archive for later iterations.

\section{Implementation Details}
\label{app:impl_details}

In this section, we provide more implementation details.

\subsection{LLM Inference}
\label{app:llm_config}

We use the following generation parameters for all models
unless otherwise noted:
temperature $= 0.2$, top-$p = 0.95$, and maximum output tokens $= 4{,}096$.
For reasoning-capable models, we increase the maximum output tokens to 16{,}384
to accommodate extended chain-of-thought generation.

For proprietary models (GPT-5, GPT-5.4-mini, Claude 3.5 Sonnet),
we use either the Snowflake Cortex LLM inference service
or the respective provider's API (OpenAI, Anthropic).
For open-source models (Qwen3-8B),
we serve them locally via Ollama on a server equipped with
8$\times$ NVIDIA RTX 3090 GPUs (24\,GB each).
All LLM calls are routed through a unified LiteLLM proxy endpoint
to ensure a consistent interface across providers.

\subsection{Database Interaction}
\label{app:db_interaction}

RealEHR and LS-Hard use PostgreSQL as the database backend,
while EHRSQL uses SQLite.
SQL queries are restricted to read-only operations (\texttt{SELECT}/\texttt{WITH} only);
any data-modifying statements (\texttt{INSERT}, \texttt{UPDATE}, \texttt{DELETE}, \texttt{DROP}, etc.)
are rejected before execution.
Each query is subject to a timeout of 25 seconds,
and results are capped at 10{,}000 rows.
Query results are returned to the LLM in CSV format.
On RealEHR, the tool result is truncated to 8{,}000 characters;
on LS-Hard, it is truncated to 4{,}000 characters.

For LS-Hard, the agent has access to three tools:
\texttt{execute\_sql}, \texttt{web\_search}, and \texttt{fetch\_url\_content}.
For RealEHR, only \texttt{execute\_sql} is available,
as the required groundings are institution-specific and not recoverable from the web.

\subsection{Prompt Management}
\label{app:prompt_management}

All system and user prompts are managed as Jinja2 templates,
with conditional branches that adapt to the target database engine
(PostgreSQL, SQLite, or Snowflake),
the output mode (SQL or natural-language answer),
and the available tool set.
The prompts reproduced in Appendix~\ref{app:prompts}
show the LS-Hard configuration (PostgreSQL, NLQA mode,
with \texttt{web\_search} enabled);
other configurations differ primarily in SQL syntax guidance
and the final output format.

\subsection{Output Parsing and Retry}
\label{app:parse_retry}

All LLM modules are expected to return structured JSON output.
When parsing fails, the system applies an automatic retry loop
that appends the error type and a targeted hint to the conversation
and re-invokes the LLM (up to 3 attempts).
Parsing-hint messages are specialized by error type:
\texttt{end} without a valid answer,
missing or disallowed \texttt{tool\_name},
missing required tool arguments, etc.
If all retries are exhausted, the step is recorded as a parsing failure
and the exploration continues from the last valid state.
The retry prompt templates are shown in Appendix~\ref{app:prompt-parse-retry}.

\subsection{LLM-as-a-Judge Evaluation}
\label{app:llm_judge}

For both RealEHR and LS-Hard,
we evaluate the generated answers against ground-truth using an LLM judge to report answer accuracy.
We use \texttt{gpt-5.4-2026-03-05} with deterministic decoding (temperature $= 0$, top-$p = 1$) as the judge model.
Given the question, the predicted answer, and the ground-truth answer,
the judge is prompted to
(1) extract the final answer from the system's response,
(2) determine whether it matches the ground truth, and
(3) assign a binary score (1 for correct, 0 for incorrect).
For RealEHR, we additionally employ a two-pass evaluation:
an abstain classifier first determines whether the system declines to answer
(e.g., ``no records found''), and only responses classified as providing a specific answer
proceed to the correctness judge.
This allows the judge to correctly handle unanswerable questions
without conflating abstention with incorrect answers.
The RealEHR judge also applies domain-specific semantic equivalence rules
for medical terminology (abbreviations, synonyms, specificity variation).
The full prompt templates and representative input/output examples
are shown in App.~\ref{app:judge-lshard}--\ref{app:judge-examples}.

\lstdefinestyle{prompt}{%
  basicstyle=\scriptsize\ttfamily,
  frame=none,
  numbers=none,
  breaklines=true,
  breakatwhitespace=true,
  breakindent=0pt,
  breakautoindent=false,
  showstringspaces=false,
  aboveskip=3pt,
  belowskip=0pt,
  xleftmargin=0pt,
  columns=fullflexible,
  keepspaces=true,
}

\subsubsection{LS-Hard judge prompt}
\label{app:judge-lshard}

For LS-Hard, the judge receives the question, the system's response,
and the ground-truth answer, and produces a binary correctness verdict.

\begin{prompt}
Judge whether the following [response] to [question] is correct or not based on the precise and unambiguous [correct_answer] below.

[question]: {question}

[response]: {response}

Your judgement must be in the format and criteria specified below:

extracted_final_answer: The final exact answer extracted from the [response]. Put the extracted answer as 'None' if there is no exact, final answer to extract from the response.

[correct_answer]: {correct_answer}

reasoning: Explain why the extracted_final_answer is correct or incorrect based on [correct_answer], focusing only on if there are meaningful differences between [correct_answer] and the extracted_final_answer. Do not comment on any background to the problem, do not attempt to solve the problem, do not argue for any answer different than [correct_answer], focus only on whether the answers match.

correct: Answer 'yes' if extracted_final_answer matches the [correct_answer] given above, or is within a small margin of error for numerical problems. Answer 'no' otherwise, i.e. if there is any inconsistency, ambiguity, non-equivalency, or if the extracted answer is incorrect.

score: Use 1 for correct, 0 for incorrect.

confidence: The extracted confidence score between 0% and 100% from [response]. Put 100 if there is no confidence score available.
\end{prompt}

\subsubsection{RealEHR judge prompt (two-pass)}
\label{app:judge-realehr}

For RealEHR, we employ a two-pass evaluation.
In the first pass, an abstain classifier determines whether the system's response
provides a specific answer or declines to answer
(e.g., ``no records found'').
If the response is classified as a decline, the judge assigns
\texttt{correct\,=\,"abstain"} without proceeding to the second pass.
Otherwise, the second pass evaluates correctness with
domain-specific semantic equivalence rules for medical terminology.

\vspace{4pt}\noindent\textbf{Pass 1: Abstain classifier.}

\begin{prompt}
Classify whether the following [response] provides a specific answer or declines to answer the [question].

[question]: {question}

[patient_id]: {patient_id}

[department]: {department}

[response]: {response}

Classify the response:
- response_type='decline' if the response states that data cannot be found, no records exist, information is unavailable, the query returned no results, etc.
- response_type='specific_answer' if the response provides any concrete value (a date, number, drug name, diagnosis, description, etc.), even if that value might be wrong.

reasoning: Briefly explain why you classified it this way.
\end{prompt}

\vspace{4pt}\noindent\textbf{Pass 2: Correctness judge.}

\begin{prompt}
Judge whether the following [response] to [question] is correct or not based on the precise and unambiguous [correct_answer] below.

[question]: {question}

[patient_id]: {patient_id}

[department]: {department}

[response]: {response}

Your judgement must be in the format and criteria specified below:

extracted_final_answer: The final exact answer extracted from the [response]. Put the extracted answer as 'None' if there is no exact, final answer to extract from the response.

[correct_answer]: {correct_answer}

Important:
- Judge based on what the [question] explicitly asks for.
- If the [correct_answer] includes extra details not asked in the [question], do NOT penalize the [response] for omitting them.
- If the [response] includes additional correct details beyond [correct_answer], do NOT penalize.
- Treat semantically equivalent expressions as correct. For example:
  - Common abbreviations: "HTN" = "hypertension", "T2DM" = "type 2 diabetes"
  - Specificity variation: a more specific subtype that clearly refers to the same concept is acceptable (e.g., "left femoral fracture" = "femur fracture")
  - Synonym variation: different standard terms for the same concept are equivalent
- For dates: different formats of the same date are equivalent (e.g., "2020-01-15" = "Jan 15, 2020").
- For numerical values: allow small margin of error and equivalent units.
- If [correct_answer] lists multiple items, the [response] must contain ALL of them (in any order/format) to be correct.
- If the [response] mentions the correct information anywhere in its text (not just the first sentence), consider it correct.

reasoning: Explain whether the extracted_final_answer is semantically equivalent to [correct_answer]. Focus on whether the core medical facts match, not on superficial wording differences.

correct: Answer 'yes' if extracted_final_answer is semantically equivalent to [correct_answer], covering all key facts asked by [question]. Allow reasonable variation in terminology, specificity
level, and phrasing. Answer 'no' only if there is a genuinely meaningful factual difference.

score: Use 1 for correct, 0 for incorrect.

confidence: The extracted confidence score between 0% and 100% from [response]. Put 100 if there is no confidence score available.
\end{prompt}

\subsubsection{Judge input/output examples}
\label{app:judge-examples}

We show two representative judge evaluations from RealEHR
to illustrate the end-to-end evaluation pipeline.

\vspace{4pt}\noindent\textbf{Example 1: Correct answer.}

The system correctly identifies hemodialysis catheter insertion dates.

\begin{prompt}
[question]: How many times did this patient receive hemodialysis catheter insertion, and what were the dates for each?

[response]: Two hemodialysis catheter (perm catheter) insertions were found for patient R-1126-00317983.
1st: 2019-08-29 -- "Insertion of permanent catheter (right)", removal of prior non-tunneled catheter and re-insertion via Rt IJV.
2nd: 2021-09-21 -- "Insertion of permanent catheter (right)".

[correct_answer]: 2019-08-29 / 2021-09-21
\end{prompt}

\vspace{2pt}\noindent Judge output:

\begin{prompt}
extracted_final_answer: 2019-08-29 / 2021-09-21
reasoning: The response lists two catheter insertion dates: 2019-08-29 and 2021-09-21, which exactly match the correct_answer. There are no discrepancies.
correct: yes
score: 1
confidence: 100
\end{prompt}

\vspace{4pt}\noindent\textbf{Example 2: Incorrect answer.}

The system returns an ICD code rather than the expected diagnosis terms.

\begin{prompt}
[question]: What is this patient's initial diagnosis?

[response]: Initial diagnosis: M348 (systemic sclerosis with pulmonary involvement).

[correct_answer]: diffuse systemic sclerosis, interstitial lung disease
\end{prompt}

\vspace{2pt}\noindent Judge output:

\begin{prompt}
extracted_final_answer: M348 (systemic sclerosis with pulmonary involvement)
reasoning: The response indicates systemic sclerosis with pulmonary involvement using ICD code M34.8, but it does not specify the subtype as "diffuse". The correct answer requires "diffuse systemic sclerosis" plus "interstitial lung disease". The lack of "diffuse" makes the response non-equivalent to the correct answer.
correct: no
score: 0
confidence: 100
\end{prompt}

\subsection{LLM Answer Selector}
\label{app:llm_answer_selector}

Both \ours and the repeated-ReAct baseline (\S\ref{subsec:cost_analysis})
produce multiple candidate answers per question.
A shared LLM-based answer selector chooses the single best answer
from the candidate pool, ensuring a fair comparison.

The selector receives the original question and a deduplicated list of candidates.
Candidates that are identical (after case-folding and truncation to 200 characters) are merged,
and the selector is shown the \emph{frequency} (how many runs produced the same answer)
as an additional signal.
Up to 10 unique candidates are presented.

The selector uses \texttt{gpt-5.4-mini-2026-03-17} with temperature $= 0.2$ and top-$p = 0.95$.

\paragraph{Prompt.}
The system message is:

\begin{prompt}[fontsize=\small]
You are an expert answer selector. Pick the best candidate answer by number.
\end{prompt}

\noindent The user message follows this template:

\begin{prompt}[fontsize=\small]
Question: {question}
Below are {N} candidate answers:
{[1]} (found by {freq} path(s), depth={depth})
Answer: {answer}
Evidence: {evidence}
{[2]} \ldots
Select the BEST answer by responding with ONLY its number.
Guidelines:
- Prefer answers backed by stronger evidence (SQL query results over inferences).
- Higher frequency (found by more paths) is a positive signal but not decisive.
- A more complete answer (superset) is preferred over a partial one.
- If NONE of the candidates are reliable enough, respond with 0 to abstain.
Respond with a single number (1--{N}, or 0 to abstain).
\end{prompt}

\noindent
The selector returns a single integer.
If it returns 0, the system abstains for that question.
If parsing fails after one attempt, the system falls back to the first candidate answer.

For the repeated-ReAct baseline, the \texttt{depth} field is not available and is shown as ``?'';
the \texttt{evidence} field is omitted.
For GATE, both fields are populated from the trajectory archive: depth is the number of grounding-update steps in the trajectory, and evidence is the summary of SQL checks and observations supporting the induced partially grounded plan.

\subsection{Benchmark Details}
\label{app:benchmark_details}

This section collects additional details for each benchmark.
RealEHR tables include prescriptions, diagnoses, and laboratory results in which a single concept (e.g., a drug or disease) appears under many surface forms: trade names, generic names, Korean translations, hospital-specific codes, and clinical abbreviations (examples in App.~\ref{app:realehr_examples}).
EHRSQL introduces grounding challenges through question paraphrasing, time-format mixing, and temporal data shifting; we use a stratified 100-question subset (App.~\ref{app:ehrsql_subset}).
LS-Hard removes web-recoverable grounding cues from LiveSQLBench; 180 questions satisfy the web-recoverability constraint and output short natural-language answers.

\subsubsection{Schema complexity}
\label{app:benchmark_schema}

Table~\ref{tab:benchmark_schema} compares the schema characteristics of the three evaluation benchmarks.
EHRSQL offers a moderate number of narrow tables on a single SQLite database,
LS-Hard distributes queries across 18 independent PostgreSQL databases with wider tables,
and RealEHR presents the most challenging schema with 29 heavily columned tables (up to 73 columns each) whose names and values mix Korean and English.

\begin{table}[t]
\centering
\small
\setlength{\tabcolsep}{1pt}
\begin{tabular}{l ccc}
    \toprule
    \textbf{Metric} 
    & \textbf{EHRSQL} 
    & \textbf{LS-Hard} 
    & \textbf{RealEHR} \\
    \midrule
    \textbf{Engine}      & SQLite     & PostgreSQL & PostgreSQL \\
    \textbf{Tables/DB}   & 17         & 9.7        & 29         \\
    \textbf{Cols/Table}  & 6.5        & 13.1       & 37.6       \\
    \textbf{Rows/Table}  & 18.4K      & 1.0K       & $>$100K    \\
    \bottomrule
\end{tabular}
\caption{
Schema complexity comparison across benchmarks.
EHRSQL uses a single MIMIC-IV database (312K total rows).
LS-Hard spans 18 independent databases with DDL definitions averaging 17K characters.
RealEHR operates on a proprietary clinical data warehouse (CDW) with mixed Korean--English column names and up to 73 columns per table; exact row counts cannot be disclosed.
}
\label{tab:benchmark_schema}
\end{table}

\begin{table*}[t]
\centering
\begin{tabular}{@{}llrrrc@{}}
\toprule
\textbf{Dimension} & \textbf{Category} & \textbf{Full (\%)} & \textbf{Sub (\%)} & \textbf{$\Delta$} & \textbf{MAD} \\
\midrule
\multirow{2}{*}{Answerability}
  & Answerable    & 80.1 & 80.0 & $-$0.1 & \multirow{2}{*}{0.1} \\
  & Unanswerable  & 19.9 & 20.0 & $+$0.1 & \\
\midrule
\multirow{3}{*}{SQL Complexity}
  & Simple   &  2.4 &  3.0 & $+$0.6 & \multirow{3}{*}{0.9} \\
  & Medium   & 55.9 & 57.0 & $+$1.1 & \\
  & Complex  & 21.8 & 20.0 & $-$1.8 & \\
\midrule
\multirow{5}{*}{Primary Table (top 5)}
  & \texttt{admissions}       & 11.0 & 11.0 &  0.0 & \multirow{5}{*}{0.3} \\
  & \texttt{prescriptions}    & 10.8 & 11.0 & $+$0.2 & \\
  & \texttt{labevents}        &  7.8 &  7.0 & $-$0.8 & \\
  & \texttt{chartevents}      &  7.2 &  7.0 & $-$0.2 & \\
  & \texttt{diagnoses\_icd}   &  7.2 &  7.0 & $-$0.2 & \\
\midrule
Question Template
  & \multicolumn{3}{l}{\textit{135 unique; 68 in subset}}
  & & 0.6 \\
\bottomrule
\end{tabular}
\caption{Distribution comparison between the full EHRSQL validation set ($n{=}1{,}163$) and the stratified subset ($n{=}100$). $\Delta$ denotes the percentage-point difference (subset $-$ full). MAD is the mean absolute deviation across categories within each dimension.}
\label{tab:ehrsql_subset_dist}
\end{table*}

\subsubsection{EHRSQL subset sampling}
\label{app:ehrsql_subset}

The full EHRSQL validation set contains 1{,}163 questions (931 answerable, 232 unanswerable).
To keep experimental costs manageable while preserving the distribution of the original set,
we construct a balanced 100-question subset via stratified sampling.
The subset comprises 80 answerable and 20 unanswerable questions,
maintaining the approximate 80/20 ratio of the full set.
Stratification is performed along two dimensions:
(1) the \emph{primary table} referenced in the ground-truth SQL
and (2) \emph{SQL complexity} (simple single-table queries vs.\ multi-table joins and aggregations).
This ensures that all 17 MIMIC-IV tables are represented
and that the complexity distribution mirrors the full validation set.

Table~\ref{tab:ehrsql_subset_dist} reports the distribution match
across four stratification dimensions.
The mean absolute deviation (MAD) between the full-set and subset proportions
is below 1\,pp on every dimension,
confirming that the 100-question subset faithfully represents
the original validation set.

\subsubsection{RealEHR example questions}
\label{app:realehr_examples}

RealEHR is a closed benchmark built on a proprietary clinical data warehouse; the underlying data cannot be released.
To give readers a sense of the task, Table~\ref{tab:realehr_examples} presents two representative questions from the 47-item evaluation set (both medium difficulty).
Questions are translated from Korean and patient identifiers are anonymized.
The original questions use the informal shorthand of routine chart review, while the underlying database stores concepts under heterogeneous surface forms (trade names, generic names, Korean translations, hospital codes, and clinical abbreviations), creating a naturally occurring semantic gap.

\subsection{Run Variance}
\label{app:run_variance}

Table~\ref{tab:main_results_std} reports the mean performance and standard deviation across independent runs.
We ran each setting multiple times: the ReAct setting on RealEHR averages 6 runs, while all other values are averaged across 5 runs.

\begin{table*}[t]
\centering
\setlength{\tabcolsep}{6pt}
\begin{tabular}{llccc}
    \toprule
    \textbf{Backbone} & \textbf{Method}
    & \textbf{RealEHR} & \textbf{EHRSQL} & \textbf{LS-Hard} \\
    \midrule
    \multirow{4}{*}{Qwen3-8B}
    & ReAct~\citep{yao_react_2023}
      & 1.7 \( \pm {\scriptstyle 1.8}\)
      & 30.2 \( \pm {\scriptstyle 1.3}\)
      & 14.5 \( \pm {\scriptstyle 1.1}\) \\
    & ReDel~\citep{zhu2024redel}
      & 2.5 \( \pm {\scriptstyle 1.0}\)
      & 26.4 \( \pm {\scriptstyle 2.4}\)
      & 8.0 \( \pm {\scriptstyle 1.2}\) \\
    & ReFoRCE~\citep{deng_reforce_2025}
      & 3.8 \( \pm {\scriptstyle 1.8}\)
      & 31.2 \( \pm {\scriptstyle 1.1 }\)
      & 13.9 \( \pm {\scriptstyle 0.6 }\) \\
    & \textbf{\ours}
      & \(\textbf{5.6} \pm {\scriptstyle 1.2 }\)
      & \(\textbf{35.0} \pm {\scriptstyle 1.0 }\)
      & \(\textbf{17.3} \pm {\scriptstyle 0.9}\) \\
    \midrule
    \multirow{4}{*}{GPT-5.4-mini}
    & ReAct~\citep{yao_react_2023}
      & 42.9 \( \pm {\scriptstyle 6.8}\)
      & 32.2 \( \pm {\scriptstyle 1.5}\)
      & 37.8 \( \pm {\scriptstyle 2.4}\) \\
    & ReDel~\citep{zhu2024redel}
      & 44.8 \( \pm {\scriptstyle 1.9}\)
      & 26.8 \( \pm {\scriptstyle 1.3}\)
      & 40.2 \( \pm {\scriptstyle 2.2}\) \\
    & ReFoRCE~\citep{deng_reforce_2025}
      & 48.6 \( \pm {\scriptstyle 0.7}\)
      & 27.6 \( \pm {\scriptstyle 0.9}\)
      & 40.1 \( \pm {\scriptstyle 1.1}\) \\
    & \textbf{\ours}
      & \(\textbf{55.2} \pm {\scriptstyle 1.0}\)
      & \(\textbf{34.4} \pm {\scriptstyle 1.1}\)
      & \(\textbf{45.1} \pm {\scriptstyle 0.7}\) \\
    \bottomrule
\end{tabular}
\caption{
Mean performance and standard deviation across independent runs.
RealEHR and LS-Hard report answer accuracy (\%); EHRSQL reports execution accuracy (\%).
}
\label{tab:main_results_std}
\end{table*}

\begin{table*}[ht]
\centering
\small
\setlength{\tabcolsep}{5pt}
\begin{tabular}{p{1.6cm} p{\dimexpr\textwidth-1.6cm-4\tabcolsep-2\arrayrulewidth}}
    \toprule
    \multicolumn{2}{l}{\textbf{Example 1} — Cardiology, CHA\textsubscript{2}DS\textsubscript{2}-VASc score calculation} \\
    \midrule
    Question & For a patient with atrial fibrillation, what is the CHA\textsubscript{2}DS\textsubscript{2}-VASc score? \\
    Answer & 1 \\
    \addlinespace
    DB tables & \texttt{FT\_CRE\_DGNS} (diagnoses), \texttt{FT\_CRE\_VIST} (visits) \\
    Key columns & \texttt{CLDG\_VOC\_NM} (diagnosis name), \texttt{ICD10\_CD} (ICD-10 code), \texttt{AGE\_CD\_YY} (age), \texttt{SEX\_TP\_NM} (sex) \\
    \addlinespace
    DB output & \texttt{Atrial fibrillation | 69 | male | 2023-08-25} \\
              & \texttt{Atrial fibrillation | 70 | male | 2024-04-15} \\
    \addlinespace
    Gap & The agent must (1) know that CHA\textsubscript{2}DS\textsubscript{2}-VASc is a stroke risk score computed from comorbidities, age, and sex; (2) identify the relevant ICD-10 code families (\texttt{I\%}, \texttt{E08--E13\%}); and (3) apply the scoring rule to the retrieved values (age 65--74 $=$ 1\,pt for a male patient with no other qualifying diagnoses). \\
    \midrule
    \addlinespace
    \multicolumn{2}{l}{\textbf{Example 2} — Pulmonology, QT-prolonging medication identification} \\
    \midrule
    Question & Which medications could have caused the prolonged QT interval? \\
    Answer & Azithromycin, clofazimine \\
    \addlinespace
    DB tables & \texttt{FT\_CRE\_MDPR} (prescriptions), \texttt{FT\_CRE\_EXM\_FUN} (functional exams) \\
    Key columns & \texttt{INGR\_MDPR\_NM} (ingredient/drug name), \texttt{MDFM\_CPEM\_NM} (exam component name), \texttt{ORD\_DT} (order date) \\
    \addlinespace
    DB output & \texttt{Amikacin | 2023-10-04 | 2023-11-23} \\
              & \texttt{Azithromycin | 2023-10-04 | 2024-02-22} \\
              & \texttt{Clofazimine | 2023-10-04 | 2024-03-21} \\
              & \texttt{Esomeprazole | 2023-10-04 | 2023-10-30} \\
              & \texttt{Ethambutol | 2023-10-04 | 2024-03-21} \\
              & \texttt{Metoclopramide | 2023-10-05 | 2023-11-08} \\
    \addlinespace
    Gap & The agent must (1) discover that \texttt{MDFM\_CPEM\_NM} stores QTc measurements by searching for the string \texttt{`QTc'}; (2) join prescriptions temporally with the abnormal QTc finding; and (3) apply pharmacological knowledge to filter the six co-administered drugs down to the two known QT-prolonging agents. \\
    \bottomrule
\end{tabular}
\caption{
Two representative medium-difficulty questions from the RealEHR benchmark.
Patient identifiers are anonymized and questions are translated from Korean.
``DB tables'' and ``Key columns'' show the hospital-internal naming conventions the agent must navigate; ``Gap'' summarizes the semantic reasoning required beyond SQL generation.
}
\label{tab:realehr_examples}
\end{table*}

\section{Prompt Templates}
\label{app:prompts}

We reproduce the core system and user prompts for each method below
(see Appendix~\ref{app:prompt_management} for how templates are configured).

ReAct uses the base ReAct agent shown in App.~\ref{app:prompt-react}.
ReDel builds a coordinator/child-agent architecture on top of that agent (App.~\ref{app:prompt-redel}).
\ours reuses the same agent as a submodule for executing SQL checks from selected trajectories, and adds prompts for trajectory evaluation (App.~\ref{app:trajectory_evaluation_prompt}) and evidence summarization for repeated execution (App.~\ref{app:evidence_summarization}).
Output parsing and retry prompts are shared across all methods (App.~\ref{app:prompt-parse-retry}).

The prompts use implementation-level terminology (e.g., \texttt{dist\_total} for progress, \texttt{d\_potential} for potential, \texttt{lineage\_density} for the UCB exploration term) that differs from the abstractions introduced in \S\ref{sec:proposed_method}, but each prompt component implements the same functional role described in the main text.

\lstdefinestyle{prompt}{%
  basicstyle=\scriptsize\ttfamily,
  frame=none,
  numbers=none,
  breaklines=true,
  showstringspaces=false,
  aboveskip=3pt,
  belowskip=0pt,
  xleftmargin=6pt,
  columns=fullflexible,
}

\subsection{ReAct Agent}
\label{app:prompt-react}

The ReAct agent is the base tool-calling module that interacts with external tools via sequential tool calls
(\texttt{execute\_sql}, \texttt{web\_search}, \texttt{fetch\_url\_content}).
It is used directly as the sole agent in ReAct, as the root and child agents in ReDel (with additional coordination logic), and as a submodule for proposing and executing SQL checks over grounded subplans in \ours.

\vspace{4pt}\noindent\textbf{System message.}

\begin{prompt}
## SYSTEM

You are a ReAct agent for database question answering.

## Methodology: Linear/Sequential

- Single agent executes tools sequentially (no recursion or delegation).
- Budget limit: {tool_call_limit} tool calls maximum.
- Schema discovery first: discover the database schema by querying information_schema.tables (and then columns for relevant tables). Never guess or assume table/column names.

## Role & Decisions

- Use tool_call to gather information or verify SQL.
- If the question requires unfamiliar domain-specific knowledge, strongly prefer using web_search to retrieve it before finalizing.
- A single successful SQL execution is NOT sufficient to finalize. Always validate with cross-checks (alternate aggregation, sanity checks on row counts/nulls, or spot-checks) before ending.
- End with next_action="end" and provide the final answer.
- Prefer multiple small, independent SQL checks over a single monolithic query.

## Common Mistakes to Avoid

- Returning untested SQL
- Drafting SQL before confirming schema/columns via tools
- Relying on a single query without cross-checks
- Ignoring conflicting evidence between queries

## Actions

1. Tool Call:  {"next_action": "tool_call", "tool_name": ..., "tool_kwargs": ...}
2. Final Answer: {"next_action": "end", "answer": "..."}
\end{prompt}

\vspace{4pt}\noindent\textbf{User message.}

The user message contains the question, external knowledge (if any),
the database schema (DDL with sample rows), and column meaning annotations.
In multi-turn settings (ReDel children, \ours rollouts),
the conversation history of prior tool calls and results is appended.

\begin{prompt}
Question: {question_text}

Utilize the following information if needed.
{external_knowledge}

Database ID: {db_id}

# Database Schema (Local)
CREATE TABLE "sprint_results" (
    srescode integer NOT NULL,
    matchref integer NOT NULL,
    ...
    sprint_performance jsonb NULL,
    PRIMARY KEY (srescode)
);

First 3 rows:
  srescode  matchref  ...  sprint_performance
  --------  --------  ...  ---------------------------------------------------
         5      1061  ...  {'grid': 6, ...
...

# Column Meanings (Local)
{column_meaning_annotations}
\end{prompt}

\subsection{ReDel (Multi-Agent)}
\label{app:prompt-redel}

ReDel~\citep{zhu2024redel} reuses the base ReAct agent from App.~\ref{app:prompt-react}, but augments it with a coordinator/child-agent architecture.
The \textbf{coordinator} (root agent) works on the original question and may delegate narrower investigation subtasks to \textbf{child agents} via \texttt{spawn\_child}.
Child agents investigate their assigned subtasks and return their findings to the coordinator, which integrates them and produces the final answer.
We show only the additional ReDel-specific instructions below; the shared SQL guidance and tool descriptions from App.~\ref{app:prompt-react} are omitted.

\vspace{4pt}\noindent\textbf{System message (coordinator).}

\begin{prompt}
## SYSTEM

You are a Coordinator Agent for database question answering with
exploration agents.

## Approach

- You solve the problem directly. Use your own tool calls for schema exploration, querying, and SQL verification.
- You have access to exploration agents (spawn_child) for delegating narrow investigation tasks, but use them only as a last resort.
- Only YOU produce the final answer. Children only report findings.

### When to Spawn

- Before your first action, estimate how many tool calls this task needs.
- Do NOT spawn if you can finish within your remaining budget. Most tasks need 5-15 tool calls -- handle them yourself.
- Spawn ONLY when: you have exhausted >60%
- Never spawn on the first few turns. Always start by working directly.

### If You Must Spawn

- Give specific, narrow subtasks such as:
  - "List all tables in schema X and describe their columns"
  - "Check if table Y contains records matching condition Z"
  - "Verify the join between tables A and B on column C"
- Bad subtasks (too broad): "Solve the query", "Find the answer"
- After receiving child findings, integrate them and continue working.

## Actions

1. Tool Call:  {"next_action": "tool_call", ...}
2. Spawn Child: {"next_action": "spawn_child", "subtask": "..."}
3. Final Answer: {"next_action": "end", "answer": "..."}
\end{prompt}

\vspace{4pt}\noindent\textbf{System message (child agent).}

\begin{prompt}
## SYSTEM

You are an Exploration Agent for database investigation.

## Approach

- You are assigned a focused exploration subtask. Your job is to investigate and report findings -- NOT to answer the main question.
- Use SQL and other tools to discover schema structure, verify data values, and test query patterns.
- Report your findings as structured facts: what tables exist, what columns are relevant, what values you found, what constraints apply.
- Do NOT attempt to produce a final answer for the main question.

## Actions

1. Tool Call:  {"next_action": "tool_call", ...}
2. Report Findings: {"next_action": "end", "answer": "{structured findings}"}
\end{prompt}

\vspace{4pt}\noindent\textbf{User message (coordinator).}

Same as the ReAct agent (App.~\ref{app:prompt-react}),
with additional dynamic context:

\begin{prompt}
{same question, schema, and column meanings as ReAct}

## DYNAMIC CONTEXT (AUTO-GENERATED)

# Role
- Root agent (depth 0/1)

# Database Context
- Target DB Engine: postgresql
- Target Database ID: {db_id}

# Budgets
- Agent: {used}/{limit} tool calls
- Global: {global_used}/{global_limit} tool calls
\end{prompt}

\vspace{4pt}\noindent\textbf{User message (child agent).}

\begin{prompt}
{same schema and column meanings as coordinator}

## DYNAMIC CONTEXT (AUTO-GENERATED)

# Role
- Child agent (depth 1/1)
- Subtask: {subtask_description}

# Budgets
- Agent: {used}/{child_limit} tool calls
- Global: {global_used}/{global_limit} tool calls
\end{prompt}

\subsection{\ours (Ours)}
\label{app:prompt-cave}

\ours reuses the ReAct agent from App.~\ref{app:prompt-react} to execute SQL checks from a selected trajectory-induced plan.
Its additional modules implement the search and update loop in Algorithm~\ref{alg:ours}.
The trajectory evaluator selects which trajectory-induced plan to continue.
The local evidence summarizer converts executed SQL checks and observations into update summaries \(u\).
When such a summary grounds an operator, the corresponding execution step \(a=(\tilde{x},o,u)\) is stored as a reusable grounding update and may be appended to trajectories in later iterations.

The prompts below use implementation-level names inherited from the codebase.
In the terminology of Section~\ref{sec:proposed_method}, a ``state'' corresponds to a trajectory-induced partially grounded plan, and an ``action'' corresponds to an execution step.
We retain the raw prompt wording for reproducibility.

\subsubsection{Trajectory evaluation (\(f_{\mathrm{LLM}}\))}
\label{app:trajectory_evaluation_prompt}

The trajectory evaluator receives the query \(q\), a candidate trajectory \(T_i\), the current trajectory archive \(\mathbb{T}_t\), and the update archive \(\mathbb{U}_t\).
It evaluates the partially grounded plan induced by \(T_i\) along three axes:
\emph{progress}, measuring how much of the target SQL is already grounded;
\emph{potential}, estimating whether the current bottleneck can lead to useful progress; and
\emph{bottleneck}, identifying the ungrounded operator whose grounding should be tested next.
This implements the plan evaluation step described in \S\ref{subsec:iterative_search}.

The raw prompt below uses the implementation term ``state'' for a trajectory-induced plan and ``frontier'' for the current trajectory archive; potential is expressed as \texttt{d\_potential}, the estimated remaining tool calls to complete.

\vspace{4pt}\noindent\textbf{System message.}

\begin{prompt}
You are the state evaluator for tree-search exploration.

## Your Role

You receive a snapshot of the current state frontier. Each state has:
- dist_total: Running estimate of remaining distance to a correct answer
- lineage_selected: Cumulative selection count (own + all ancestors)
- action history: What the trajectory has done so far (by action IDs)
- endpoint: Summary of the trajectory's current state

Your task: annotate each state with d_potential -- a score combining:
1. Optimistic remaining distance if its primary bottleneck is resolved
2. Exploration incentive for under-explored or novel states

## Understanding dist_total

Updated after every tool call based on concrete outcomes:
- Decreases on progress: SQL returning useful rows (~-1.0), non-SQL tool (~-0.15)
- Increases on setback: SQL error (~+1.0), empty result (~+0.4)
- Lower dist_total = closer to a correct answer

## Scale Definition (First Call)

d_potential for the initial state is an absolute estimate of tool calls needed
from scratch -- it defines the distance unit for the entire run.

- Simple (single table, direct lookup): 2-4
- Medium (multi-table join, aggregation): 5-10
- Complex (exploratory schema, unclear structure): 10-20

## Bottleneck & Potential Value

d_potential estimates "optimistic remaining distance" -- how close the state
could be IF the bottleneck were resolved.

CRITICAL: d_potential ~ 0-1 REQUIRES ALL of the following:
1. At least two independent SQL queries returned the same answer
2. The state's knowledge explicitly notes cross-verification
3. No alternate candidate values exist

Minimum d_potential for unverified single-query answers: 2.0

## UCB-like Exploration Bonus

Use lineage_density = lineage_selected / max(1, depth):
- lineage_density <= 1.5: fresh region -> exploration bonus (reduce d_potential by 1-3)
- lineage_density 1.5-2.5: no adjustment
- lineage_density > 2.5 + dist_total stagnant: over-explored -> penalty (+2-5)
- lineage_density > 2.5 + dist_total decreasing: progressing -> no penalty

## Reasoning Process (per state)

1. Review action path and endpoint state
2. Compare dist_total with depth -- is distance decreasing?
3. Compute lineage_density. Over-explored or fresh?
4. Identify the primary bottleneck
5. Check cross-verification status:
   - 0 confirmations (schema only): d_potential >= 8
   - 1 SQL result, unverified: d_potential 4-7
   - 2+ independent results agree: d_potential 2-4
   - Fully cross-verified: d_potential 0-2
6. Estimate remaining tool calls; apply exploration adjustment
7. When in doubt, assign HIGHER d_potential (optimism under uncertainty)

## Output

Return JSON: {analysis_summary, state_values: [{trajectory_id, d_potential,
bottleneck_note}]}
\end{prompt}

\vspace{4pt}\noindent\textbf{User message.}

\begin{prompt}
## Action Dictionary
[a1] {action_summary_1}
[a2] {action_summary_2}
...

## Frontier Trajectories (top {frontier_k})
[t0] depth=0, dist_total=60.0, lineage_selected=0
  path: (none)
  endpoint: (initial state)
  last_result: (none)
  knowledge: (none)
  sql_success: N/A

[t1] depth=2, dist_total=8.0, lineage_selected=3
  path: a1 -> a2
  endpoint: {endpoint_summary}
  last_result: {last_tool_result_summary}
  knowledge: {knowledge_snapshot}
  sql_success: {sql_success_rate}
...

## Output Constraint
You MUST output state_values only for trajectory_ids in this list:
['t0', 't1', ...]

Remember:
- For t0 (no actions yet): provide your absolute estimate of total tool calls
  needed from scratch. This value becomes the distance unit for the entire run.
- lineage_selected: total times this trajectory AND all ancestors have been
  selected for expansion -- use for UCB adjustment.
- Lower d_potential = closer to solution after bottleneck fix = higher potential
\end{prompt}

\subsubsection{Evidence summarization for repeated execution}
\label{app:evidence_summarization}

As described in \S\ref{subsec:partial_execution}, one SQL check may only partially resolve the grounding of an ungrounded operator.
When \ours performs another round of execution for the same operator, it conditions the next SQL check on what previous checks have established, refined, or ruled out.
The evidence summarization prompt implements this step.
Given the executed SQL checks, their observations, and the recent trajectory context, it produces structured summaries that identify useful evidence, failed interpretations, and remaining uncertainty for the current bottleneck operator.

These summaries serve two roles.
First, per-action summaries provide the update summary \(u\) in an execution step \(a=(\tilde{x},o,u)\).
When \(u\) grounds an operator, the corresponding step is stored in the update archive and can be appended to trajectories.
Second, local guidance fields such as \texttt{critical\_advantages} and \texttt{avoid\_actions} guide subsequent execution rounds for the same selected trajectory; they are used to target the remaining uncertainty but are not themselves promoted to reusable grounding updates unless they support an operator grounding.

The raw prompt below uses implementation field names such as \texttt{action\_summaries}, \texttt{local\_summary}, \texttt{critical\_advantages}, and \texttt{avoid\_actions}.
These fields implement the repeated-execution mechanism described above.

\vspace{4pt}\noindent\textbf{System message.}

\begin{prompt}
You are the combined Action Summarizer + Local Reflection module,
operating in BATCHED mode.

You receive a batch of raw actions (decision + tool result pairs) from a
single rollout, plus local attempt context. You must produce BOTH:
1. Per-action summaries with progress_delta and knowledge_updates
2. Local reflection (local_summary + critical_advantages)

## Part 1: Action Summaries

For each pending action, produce:
- action_summary: 1-2 sentence factual summary of what the tool call achieved.
- progress_delta: How much closer (negative) or further (positive) this action
  moved toward a correct answer.

    Outcome                                       Delta
    SQL returns useful rows (confirms approach)    ~ -2.0
    SQL success but partial / minor data           ~ -1.5
    Schema / metadata fetch, non-SQL tool          ~ -1.0
    Empty result (first time)                      ~ -0.5
    SQL error (first time this approach)           ~  0.0
    Empty/error repeating a known mistake          ~ +1.0

- knowledge_updates: list of {operation: "add"|"edit", knowledge_id, statement}
  Write only obtained facts. No process narration or next-step plans.

## Part 2: Local Reflection

- Focus on 1-4 critical states (schema/table choice, join key selection,
  filter interpretation, aggregation/grouping, verification stage).
- Compare WHOLE attempts, not only the final action.
- Extract actionable, causal lessons (what works, what to avoid, why).

### Guiding the Next Rollout

The next rollout does NOT see global memory or tool results from other branches.
The ONLY cross-branch information it receives is action_advantages and
avoid_actions. Therefore:

- In action_advantages, include concrete schema facts:
  actual table/column names, confirmed join keys, working SQL patterns,
  data values found.
- In avoid_actions, include exact SQL patterns or column names that failed.

If previous rollouts produced terminal answers:
- State whether those answers look correct based on the SQL evidence.
- Suggest either Verify (re-derive via different path) or Diverge (previous
  answer likely wrong because [reason]).

Return strict JSON: {action_summaries: [...], local_summary, critical_advantages:
[{state_hint, action_advantages, avoid_actions}]}
\end{prompt}

\vspace{4pt}\noindent\textbf{User message.}

\begin{prompt}
## Part 1: Pending Actions to Summarize (batch mode)

### Action 0
- tool: execute_sql
- tool_kwargs: {sql: "SELECT ..."}
- decision_excerpt: {llm_reasoning_text}
- tool_result_excerpt: {tool_execution_result}

### Action 1
...

## Part 2: Local Reflection Context

### Selected Base Trajectory
- id: {base_trajectory_id}
- recent_path:
  - {action_summary_step_n-2}
  - {action_summary_step_n-1}
  - {action_summary_step_n}

- recent_steps:
  - step 1 | tool=execute_sql
    summary: {action_summary}
    result_excerpt: {tool_result_excerpt}
  - step 2 | ...

### Local Attempts from Same Base
- id: t13
  depth: 3
  last_action_summary: {summary}
  recent_path: ...
  recent_steps: ...

- id: t14
  ...

### Terminal Answers from Previous Rollouts
- t13 (depth=3): 0

When writing critical_advantages, include guidance on whether to verify or
diverge from previous terminal answers.

### Allowed Local State Hints (must use only these)
- {state_hint_1}
- {state_hint_2}

Produce JSON with:
- action_summaries: [{action_index, action_summary, progress_delta,
  knowledge_updates}]
- local_summary: include whether previous terminal answers should be trusted
- critical_advantages: [{state_hint, action_advantages, avoid_actions}]
\end{prompt}

\subsection{Output Parsing and Retry}
\label{app:prompt-parse-retry}

When the LLM produces output that cannot be parsed into valid JSON,
the system appends one of the following messages to the conversation
and re-invokes the LLM.
These templates are shared across all methods.

\vspace{4pt}\noindent\textbf{Generic retry (on JSON parse failure).}

\begin{prompt}
Parsing failed (attempt {attempt}/{max_attempts}).
Error: {error_type}: {error_message}
Before the JSON block, write brief plain-text reasoning (1-3 sentences).
Fix the JSON output and return a single ```json code block.
Do not add text after the JSON block or repeat this message.
\end{prompt}

\vspace{4pt}\noindent\textbf{Format reminder (appended alongside retry).}

\begin{prompt}
Before the JSON block, write brief plain-text reasoning (1-3 sentences).
Return one JSON code block matching this schema:
{format_instructions}
Do not include extra text after the JSON block or repeat this reminder.
\end{prompt}

\vspace{4pt}\noindent\textbf{Hint: \texttt{end} without answer.}

\begin{prompt}
[Parsing Hint]
You set next_action to "end" but did not provide a valid answer.
Either:
- Provide a non-empty "answer" (and "sql" if required), or
- Choose next_action "tool_call" instead.
Return a corrected JSON response only.
\end{prompt}

\vspace{4pt}\noindent\textbf{Hint: missing or disallowed tool name.}

\begin{prompt}
[Parsing Hint]
You set next_action to "tool_call" but did not provide a valid tool_name.
Choose one of the allowed tools: {allowed_tools}.
Return a corrected JSON response only.
\end{prompt}

\vspace{4pt}\noindent\textbf{Hint: missing tool arguments.}

\begin{prompt}
[Parsing Hint]
Your tool call is missing required arguments.
Use the correct parameter for the selected tool. Examples:
- execute_sql: {"sql": "SELECT ..."}
- web_search: {"query": "..."}
Return a corrected JSON response only.
\end{prompt}

\section{Qualitative Analysis}
\label{app:qual_analysis}

We present a detailed trace comparison on the running example
from Section~\ref{sec:intro} (\texttt{sports\_events\_4} from LS-Hard)
using ReAct (linear, single-pass) and \ours.

\paragraph{Problem setup.}
The question asks to count ``Dominant Victory'' events in sprint races,
defined as races where the winner's time gap over the runner-up exceeds 5 seconds.
The key challenge is that the database stores timing information in a JSONB column
(\texttt{sprint\_performance}) with a non-obvious dual encoding:
the winner's \texttt{final\_time} is an absolute race time (e.g., \texttt{25:38.426}),
while all other positions record a \emph{relative gap} to the winner
(e.g., \texttt{+1.430} for 1.43 seconds behind).
We show the verbatim agent traces below.
The trace uses the implementation terminology introduced in App.~\ref{app:prompts};
a rollout refers to a single grounding attempt from a selected trajectory.

\subsection{ReAct Agent (GPT-5)}

\lstdefinestyle{sqltrace}{%
  language=SQL,
  basicstyle=\scriptsize\ttfamily,
  keywordstyle=\bfseries,
  frame=none,
  numbers=none,
  breaklines=true,
  showstringspaces=false,
  aboveskip=3pt,
  belowskip=0pt,
  xleftmargin=10pt,
  columns=fullflexible,
  morekeywords={ILIKE,JSONB,WITH,FILTER,PARTITION,OVER},
  literate={->}{{\texttt{->}}}2 {->>}{{\texttt{->{}>}}}3,
}

\newcommand{\agentlabel}[1]{\vspace{6pt}\noindent\textbf{#1}\vspace{2pt}}
\newcommand{\llmout}[1]{\vspace{2pt}\noindent\textit{LLM:} #1}
\newcommand{\toolresult}[1]{\vspace{2pt}\noindent\textit{Result:} \texttt{#1}}

\agentlabel{ReAct: Tool Call 1-2} (schema \& column discovery)

\llmout{``I will first confirm the relevant tables and columns, then compute and cross-check the count of sprint races where the winner's margin over second place exceeds 5 seconds.''}

\noindent Schema discovery returns 11 tables including \texttt{sprint\_results};
column inspection reveals \texttt{sprint\_performance}\,(jsonb) as the only timing source.

\agentlabel{ReAct: Tool Call 3} (dominant victory count)

\llmout{``I'll parse \texttt{sprint\_results.sprint\_performance} to find each sprint's winner and second-place \texttt{final\_time} delta, then count events where the margin exceeds 5 seconds.''}
\begin{lstlisting}[style=sqltrace]
WITH sr AS (
  SELECT
    matchref,
    (sprint_performance->>'final_position')::int AS final_pos,
    sprint_performance->'timing'->>'final_time' AS final_time
  FROM public.sprint_results
  WHERE sprint_performance->>'final_position' IS NOT NULL
), 
per_event AS (
  SELECT
    matchref,
    MAX(CASE WHEN final_pos = 1 THEN final_time END) AS p1_final_time,
    MAX(CASE WHEN final_pos = 2 THEN final_time END) AS p2_final_time
  FROM sr
  GROUP BY matchref
), 
cleaned AS (
  SELECT
    matchref,
    p1_final_time,
    p2_final_time,
    CASE 
      WHEN p2_final_time LIKE '+%' THEN NULL -- runner-up gap string discarded!
      ELSE p2_final_time 
    END AS p2_abs_time
  FROM per_event
), 
delta_ms AS (
  SELECT
    matchref,
    ...,
    CASE 
      WHEN p2_final_time LIKE '+%' THEN NULL
      ELSE (/* parse p2_abs_time mm:ss.mmm into ms */) 
    END AS p2_ms,
    CASE 
      WHEN p1_final_time LIKE '+%' THEN (regexp_replace(p1_final_time, '^\+', ''))::numeric
      ELSE NULL 
    END AS p1_plus_s
  FROM cleaned
), 
margin AS (
  SELECT
    matchref,
    COALESCE(p1_plus_s, NULL) AS winner_gap_s,
    p2_ms
  FROM delta_ms
)
SELECT COUNT(*) AS dominant_sprint_events
FROM (
  SELECT
    matchref,
    CASE 
      WHEN winner_gap_s IS NOT NULL THEN winner_gap_s
      ELSE NULL 
    END AS gap_seconds
  FROM margin
) g
WHERE gap_seconds > 5;
\end{lstlisting}
\vspace{-2pt}
\toolresult{dominant\_sprint\_events = 0}

\noindent
The agent assumes the \textbf{winner} should carry a \texttt{+} gap prefix
(\texttt{p1\_final\_time LIKE \textquotesingle+\%\textquotesingle}),
but the winner's \texttt{final\_time} is always an absolute time (e.g., \texttt{25:38.426}).
Meanwhile, the runner-up's actual gap string (e.g., \texttt{+1.430}) is explicitly discarded
as \texttt{NULL} in the \texttt{cleaned} CTE.
As a result, \texttt{gap\_seconds} is \texttt{NULL} for every event.

\agentlabel{ReAct: Tool Call 4} (cross-check)

\llmout{``I will run a cross-check by ensuring there is at least one sprint event
and verifying parsing of the winner's gap string.''}
\begin{lstlisting}[style=sqltrace]
WITH winners AS (
  SELECT
    matchref,
    sprint_performance->'timing'->>'final_time' AS p1_time
  FROM public.sprint_results
  WHERE (sprint_performance->>'final_position')::int = 1
)
SELECT
  (
    SELECT COUNT(DISTINCT matchref) 
    FROM public.sprint_results
  ) AS total_sprint_events,
  (
    SELECT COUNT(*) 
    FROM winners
  ) AS total_winners_classified,
  (
    SELECT COUNT(*) 
    FROM winners 
    WHERE p1_time LIKE '+%'
  ) AS winners_with_gap_format,
  (
    SELECT COUNT(*) 
    FROM winners 
    WHERE p1_time ~ '^\d'
  ) AS winners_with_absolute_time_format;
\end{lstlisting}
\vspace{-2pt}
\toolresult{total\_sprint\_events=18, winners\_classified=18,
  winners\_with\_gap\_format=\colorbox{yellow!40}{0},
  winners\_with\_absolute\_time\_format=\colorbox{yellow!40}{18}}

\noindent
The cross-check confirms that \textbf{zero} winners use the \texttt{+} gap format;
all 18 have absolute times.
This is a clear signal that the gap should be read from the \textbf{runner-up's} \texttt{final\_time} instead,
but the agent does not act on it.

\agentlabel{ReAct: Final answer}

\llmout{``Winners use absolute times (not `+gap'), so gap derivation from P2 was considered;
no event met the $>$5\,s criterion.''}
\quad \texttt{answer: "0"} \quad\textcolor{red}{\ding{55}~Incorrect (ground truth: 4)}

\subsection{\ours Agent (GPT-5.4-mini)}

\agentlabel{\ours: Iteration 1, Rollout 1, Tool Call 1-2} (schema \& column discovery)

Same schema and column inspection as ReAct (omitted).
Result: identifies \texttt{sprint\_results.sprint\_performance} (JSONB) as the relevant column.

\agentlabel{\ours: Iteration 1, Rollout 2, Tool Call 3} (JSONB sample inspection)

\llmout{``I'll inspect the sprint-specific rows and their JSON timing fields to see how the winner's margin is encoded.''}
\begin{lstlisting}[style=sqltrace]
SELECT matchref, unitdrive, rno, sprint_performance
FROM   public.sprint_results
LIMIT 5;
\end{lstlisting}
\vspace{-2pt}
\textit{Result (excerpt):}\\
{\scriptsize\texttt{1061, 846, 4, \{'timing': \{'final\_time': '+24.111', 'duration\_ms': 1562537\}, 'final\_position': 5, \ldots\}}}\\
{\scriptsize\texttt{1061, 817, 3, \{'timing': \{'final\_time': '+30.959', 'duration\_ms': 1569385\}, 'final\_position': 6, \ldots\}}}

\agentlabel{\ours: Iteration 1, Rollout 2, Tool Call 4} (first aggregation --- wrong)

\llmout{``I've confirmed sprint results store time deltas in JSON, so next I need to find the winner and second-place delta for each sprint event and count those above 5 seconds.''}
\begin{lstlisting}[style=sqltrace]
WITH sprint_rows AS (
  SELECT
    sr.matchref,
    (sr.sprint_performance->>'final_position')::int AS final_position,
    sr.sprint_performance->'timing'->>'final_time' AS final_time_txt
  FROM public.sprint_results sr
  WHERE sr.sprint_performance->>'final_position' IS NOT NULL
),
parsed AS (
  SELECT
    matchref,
    final_position,
    CASE 
      WHEN final_time_txt LIKE '+%' THEN ...substring(final_time_txt from 2)::numeric
      ELSE NULL -- winner's absolute time -> NULL
    END AS delta_seconds
  FROM sprint_rows
),
per_event AS (
  SELECT
    matchref,
    MIN(delta_seconds) FILTER (WHERE final_position = 1) AS winner_delta,
    MIN(delta_seconds) FILTER (WHERE final_position = 2) AS second_delta
  FROM parsed
  GROUP BY matchref
)
SELECT COUNT(*) AS dominant_victory_events
FROM per_event
WHERE winner_delta IS NOT NULL -- always NULL (P1 has absolute time)
  AND second_delta IS NOT NULL
  AND second_delta > 5;
\end{lstlisting}
\vspace{-2pt}
\toolresult{dominant\_victory\_events = 0}
\quad\textcolor{orange}{(wrong --- same type of mistake as ReAct)}

\noindent
Like ReAct, the query fails because the winner's absolute time (\texttt{25:38.426})
has no \texttt{+}~prefix and is mapped to \texttt{NULL},
so \texttt{winner\_delta IS NOT NULL} is never satisfied.

\agentlabel{Trajectory evaluator (\(f_{\mathrm{LLM}}\)) re-analysis}

\noindent\textit{``The current aggregation only established that zero events exceed the threshold.
The main bottleneck is now parsing/validating the structured
\texttt{sprint\_performance} field and ensuring the query counts
the correct winner-vs-runner-up gap.''}

\agentlabel{Evidence summary}

\noindent\textit{``The \texttt{sprint\_results} path is semantically grounded:
winner-margin data appears in \texttt{sprint\_performance.timing.final\_time}.
However, the current $>$\,5\,s aggregation returned 0,
so the threshold or winner-vs-runner-up interpretation may be too strict or misapplied;
verify the exact gap calculation and, if needed,
test a lower threshold or an alternate parsing of the stringified structure.''}

\agentlabel{\ours: Iteration 2, Rollout 1, Tool Call 5} (key discovery)

\llmout{``I need to verify whether the previous zero came from an overly strict interpretation, so I'll inspect sprint events directly and compare winner-to-second gaps using a simpler extraction path.''}
\begin{lstlisting}[style=sqltrace]
WITH sprint_rows AS (
  SELECT
    sr.matchref,
    (sr.sprint_performance->>'final_position')::int AS final_position,
    sr.sprint_performance->'timing'->>'final_time' AS final_time_txt
  FROM public.sprint_results sr
),
ranked AS (
  SELECT
    matchref,
    final_position,
    final_time_txt,
    ROW_NUMBER() OVER (
      PARTITION BY matchref
      ORDER BY final_position
    ) AS rn
  FROM sprint_rows
  WHERE final_position IN (1, 2)
)
SELECT * FROM ranked 
ORDER BY matchref, final_position 
LIMIT 20;
\end{lstlisting}
\textit{Result (key rows):}
\begin{center}
\small
\setlength{\tabcolsep}{2pt}
\begin{tabular}{@{}rclr@{}}
\toprule
\texttt{matchref} & \texttt{pos} & \texttt{final\_time\_txt} & \\
\midrule
1061 & 1 & \colorbox{green!20}{\texttt{25:38.426}} & $\leftarrow$ absolute time \\
1061 & 2 & \colorbox{green!20}{\texttt{+1.430}} & $\leftarrow$ gap in seconds \\
1065 & 1 & \texttt{27:54.078} & \\
1065 & 2 & \texttt{+2.325} & \\
1107 & 1 & \texttt{30:26.730} & \\
1107 & 2 & \texttt{+21.048} & $\leftarrow$ gap $> 5$\,s\\
\bottomrule
\end{tabular}
\end{center}

\llmout{``I found that the winner's \texttt{final\_time} is an absolute race time,
while second place is stored as a \texttt{+} gap,
so the correct dominant-victory check is whether the second-place gap exceeds 5~seconds.''}

\agentlabel{\ours: Iteration 2, Rollout 1, Tool Call 6} (corrected query)
\begin{lstlisting}[style=sqltrace]
WITH sprint_rows AS (
  SELECT
    sr.matchref,
    (sr.sprint_performance->>'final_position')::int AS final_position,
    sr.sprint_performance->'timing'->>'final_time' AS final_time_txt
  FROM public.sprint_results sr
),
per_event AS (
  SELECT
    matchref,
    MIN(
      CASE 
        WHEN final_position = 2 AND final_time_txt LIKE '+%' THEN
          CASE 
            WHEN position(':' in substring(final_time_txt from 2)) > 0 THEN split_part(...)::numeric * 60 + split_part(...)::numeric
            ELSE substring(final_time_txt from 2)::numeric
          END
      END
    ) AS second_gap_seconds
  FROM sprint_rows
  GROUP BY matchref
)
SELECT COUNT(*) AS dominant_victory_events
FROM per_event
WHERE second_gap_seconds > 5;
\end{lstlisting}
\vspace{-2pt}
\toolresult{dominant\_victory\_events = 4}
\quad\textcolor{green!60!black}{\ding{51}~Correct}

\agentlabel{Cross-verification (multiple independent trajectories)}

\noindent $f_{\mathrm{LLM}}$: \textit{``The strongest region is the \texttt{sprint\_results} branch:
multiple independent paths now converge on the same candidate answer~4,
with the key bottleneck resolved as interpreting \texttt{timing.final\_time}
for \texttt{final\_position=2} as the relevant gap.''}

\vspace{4pt}
\noindent\textbf{Final Answer:} \texttt{4} \quad\textcolor{green!60!black}{\ding{51}}

\subsection{Discussion}

Both agents start from the same information and make the same initial mistake:
they fail to recognize that only the \emph{runner-up's} \texttt{final\_time}
encodes the gap, while the winner's is an absolute time.

The ReAct agent writes a complex query that assumes the winner carries a \texttt{+} gap prefix.
When cross-checking reveals otherwise (\texttt{winners\_with\_gap\_format = 0}), it does not revisit the interpretation and finalizes with 0.

\ours makes the same initial mistake in its first attempt, but the result is not accepted as resolved.
The trajectory evaluator \(f_{\mathrm{LLM}}\) instead identifies the remaining bottleneck as determining how \texttt{final\_time} should be interpreted for the winner and the runner-up.
A follow-up attempt then directly inspects rows with \texttt{final\_position} 1 and 2, revealing the dual encoding: the winner's \texttt{final\_time} is an absolute race time, whereas the runner-up's is a relative gap.
This observation produces the grounding update for \texttt{runner\_up\_gaps}: \texttt{runner\_up\_gap} should be parsed from rank-2 \texttt{final\_time}.
Using the grounding supported by this observation, the corrected query returns 4.

\section{Grounding-Update Conflict Case}
\label{app:eosinophil_conflict}

Section~\ref{subsec:update_reliability} reports that only one genuine conflict between valid grounding updates was found in our RealEHR audit.
In this section, we analyze this case in more detail.

\paragraph{Question.}
The question asks:

\begin{quote}
\emph{For this patient in Pulmonology, what medications and doses were administered when eosinophil was most elevated?}
\end{quote}

To answer it, the system must first ground the natural-language concept \emph{eosinophil} to a laboratory measurement in \texttt{cdw.ft\_cre\_exm\_lab}, identify the peak date or encounter, and then retrieve medications and doses from \texttt{cdw.ft\_cre\_mdpr}.
The relevant lab columns include patient identifier \texttt{pt\_no}, encounter identifier \texttt{pact\_id}, lab date \texttt{ord\_dt}, test name \texttt{exm\_nm}, and numeric result \texttt{numb\_exrs\_cnte}.
The relevant medication columns include \texttt{pt\_no}, \texttt{pact\_id}, medication date \texttt{ord\_dt}, medication names, and dosage fields.

\paragraph{Grounding A: \texttt{Eosinophil} as a percentage-like test.}
One grounding update restricts the lab test name to the exact string \texttt{Eosinophil}.
Its SQL check ranks non-null numeric results for this test and retrieves medications on the same date:
\begin{lstlisting}[style=sqlstyle, language=SQL]
WITH eos AS (
  SELECT pt_no, ord_dt, numb_exrs_cnte,
         ROW_NUMBER() OVER (
           PARTITION BY pt_no
           ORDER BY numb_exrs_cnte DESC NULLS LAST, ord_dt DESC
         ) AS rn
  FROM cdw.ft_cre_exm_lab
  WHERE pt_no = 'R-1126-00308901'
    AND exm_nm = 'Eosinophil'
    AND numb_exrs_cnte IS NOT NULL
), peak AS (
  SELECT pt_no, ord_dt AS peak_dt, numb_exrs_cnte AS peak_value
  FROM eos
  WHERE rn = 1
)
SELECT 'peak' AS src, peak_dt::text AS dt, peak_value::text AS value, NULL::text AS med
FROM peak
UNION ALL
SELECT 'med' AS src, m.ord_dt::text AS dt, NULL::text AS value,
       COALESCE(m.kor_mdpr_nm, m.eng_mdpr_nm) || ' | ' ||
       COALESCE(m.amd_qty::text, '') || ' ' || COALESCE(m.mdpr_unit_cd, '') AS med
FROM cdw.ft_cre_mdpr m
JOIN peak p ON m.pt_no = p.pt_no
WHERE m.ord_dt = p.peak_dt
ORDER BY src, dt, med;
\end{lstlisting}

The observation identifies the peak as 2024-01-28 with value 29.0000.
Under this exact-test grounding, the update is locally faithful: the SQL check supports the summary that the \texttt{Eosinophil} peak occurs on 2024-01-28, with same-day medication evidence.

\paragraph{Grounding B: \texttt{Blood Eosinophil count} as an absolute-count test.}
Other grounding updates use a broader eosinophil-name search and identify \texttt{Blood Eosinophil count} as the maximum eosinophil-related lab result:
\begin{lstlisting}[style=sqlstyle, language=SQL]
WITH eos AS (
  SELECT ord_dt, exm_nm, numb_exrs_cnte
  FROM cdw.ft_cre_exm_lab
  WHERE pt_no = 'R-1126-00308901'
    AND exm_nm ILIKE '%eosinophil%'
    AND numb_exrs_cnte IS NOT NULL
), mx AS (
  SELECT MAX(numb_exrs_cnte) AS max_val
  FROM eos
)
SELECT e.ord_dt, e.exm_nm, e.numb_exrs_cnte
FROM eos e
JOIN mx ON e.numb_exrs_cnte = mx.max_val
ORDER BY e.ord_dt, e.exm_nm;
\end{lstlisting}

The observation returns \texttt{Blood Eosinophil count} on 2023-11-18 with value 3029.0000.
A subsequent medication query retrieves medications on 2023-11-18, and a later update further restricts the evidence by the peak encounter \texttt{pact\_id}.
Under this broader eosinophil-name grounding, the update is also locally faithful: the SQL observations support 2023-11-18 as the peak date for the absolute eosinophil count.

\paragraph{Conflict from grounding ambiguity.}
The two updates do not conflict because of malformed SQL or hallucinated summaries.
Both are supported by their own SQL checks and observations.
They conflict because the natural-language phrase ``when eosinophil was most elevated'' can be grounded to two clinically related but distinct laboratory measurements:
\begin{itemize}
    \item \texttt{Eosinophil}: a percentage-like value, peaking at 29.0000 on 2024-01-28;
    \item \texttt{Blood Eosinophil count}: an absolute count, peaking at 3029.0000 on 2023-11-18.
\end{itemize}
If the task requires a single peak time, these two groundings imply different medication windows and therefore cannot both determine the same final answer.
This case therefore reflects a genuine concept-to-test-name grounding ambiguity in the clinical database.
It also explains why conflicts are rare but not impossible in RealEHR: the database can expose multiple plausible measurements for the same natural-language clinical concept.

\end{document}